\documentclass[10pt,twocolumn,letterpaper]{article}

\usepackage[pagenumbers]{wacv}

\usepackage[T1]{fontenc}
\usepackage{pifont}
\usepackage{multirow}
\usepackage{amsfonts}
\usepackage{nicefrac}
\usepackage{microtype}
\usepackage[percent]{overpic}

\definecolor{wacvblue}{rgb}{0.21,0.49,0.74}
\usepackage[pagebackref,breaklinks,colorlinks,allcolors=wacvblue]{hyperref}

\def\wacvPaperID{3399} % Replace with the paper ID assigned by CMT.
\def\confName{WACV}
\def\confYear{2027}

\title{Spatially Prompted Visual Trajectory Prediction for Egocentric Manipulation}

\author{%
  Yifan Li\textsuperscript{1}\thanks{Equal contribution. Correspondence to: \texttt{liyifa11@msu.edu}.},\quad
  Xinyu Zhou\textsuperscript{1}\footnotemark[1],\quad
  Yunhao Ge\textsuperscript{2},\quad
  Xiaobo, Tan\textsuperscript{1},\quad
  Yu Kong\textsuperscript{1}\\
  \textsuperscript{1}Michigan State University,\quad
  \textsuperscript{2}NVIDIA Research
}

\begin{document}
\maketitle

\begin{abstract}
Language instructions offer a flexible way to specify task semantics but can become ambiguous in cluttered scenes containing multiple visually similar objects and candidate targets. We study \emph{Spatially Prompted Visual Trajectory Prediction}~(SP-VTP), an open-loop setting in which points or bounding boxes in the first egocentric frame specify an object--target pair and condition the prediction of future 3D end-effector (EE) trajectory chunks. To support this setting, we introduce \emph{EgoSPT}, a controlled dataset comprising 2,841 egocentric fork pick-and-place episodes with first-frame object and target annotations and reconstructed 3D EE trajectories. SP-VTP evaluates whether a static task specification remains informative as the viewpoint and scene configuration evolve throughout execution. We also introduce \emph{SPOT}, a spatially prompted object--target trajectory predictor that integrates first-frame visual and coordinate prompts with current visual observations and motion history to generate future trajectories. Rather than relying on an explicit object tracker, SPOT maintains a fixed representation of the first-frame task specification. Across three training seeds, spatial prompting consistently reduces translation and endpoint errors relative to a no-prompt baseline, while controlled prompt interventions reveal distinct roles for object and target cues at different stages of execution. Together, EgoSPT and SPOT provide a dataset and baseline for spatially conditioned egocentric trajectory prediction in open-loop pick-and-place settings.
\end{abstract}

\section{Introduction}

\begin{figure}
    \centering
    \includegraphics[width=1\linewidth]{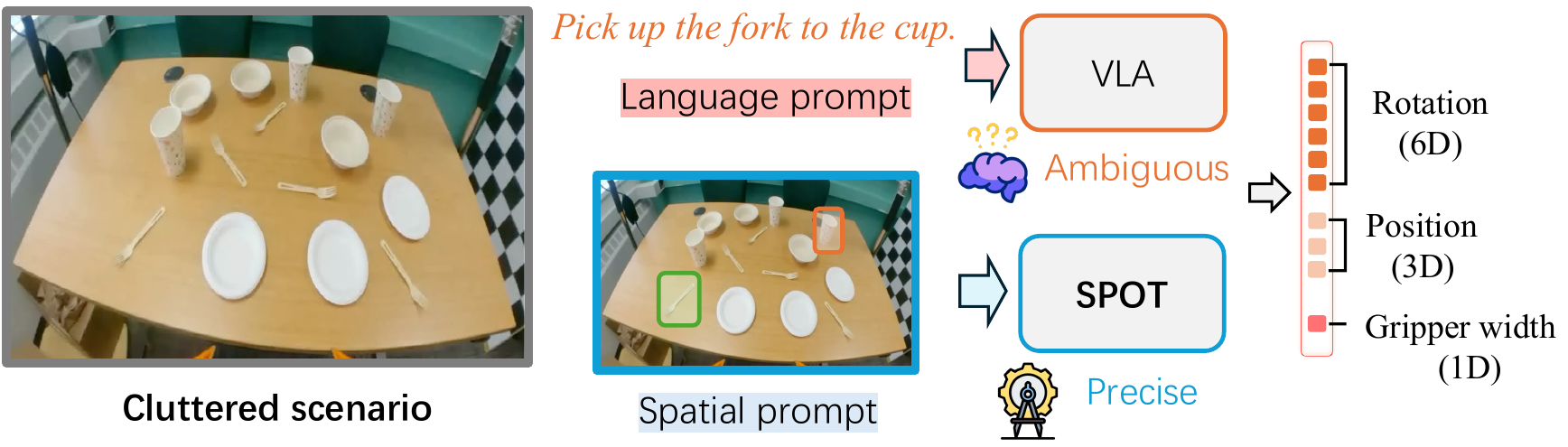}
    \caption{Motivation for SP-VTP. Language instructions can be ambiguous when multiple visually similar objects and candidate targets appear in a cluttered scene. SP-VTP uses explicit first-frame object and target prompts to condition future egocentric end-effector trajectory prediction.}
    \label{fig:motivation}
\end{figure}

% \paragraph{Manipulation as visual task specification.}
Most robot policies rely on language~\citep{lynch2021language, shridhar2022cliport, shridhar2023perceiver, zitkovich2023rt, kim2024openvla} or task identifiers~\citep{yu2020meta, yang2020multitask}. Language is expressive for task semantics, but can under-specify the intended instance or placement region in cluttered scenes with visually similar candidates. In such cases, a point or box prompt on the first egocentric frame provides a direct way to indicate \emph{this} object and \emph{that} target while preserving the visual context needed for action. We view spatial prompts as a complementary instance-grounding interface for language.

% \paragraph{A new problem: from spatial prompts to trajectories.}
We formulate this setting as \emph{Spatially Prompted Visual Trajectory Prediction} (SP-VTP). Given first-frame points or boxes indicating the object and target, the model predicts future end-effector (EE) trajectory chunks from streaming egocentric observations (see Fig.~\ref{fig:motivation}). Unlike grounding or tracking, the output is not a mask, box, or object identity, but a sequence of relative 6D EE motions and gripper states. SP-VTP is therefore an open-loop trajectory-prediction problem that tests whether a sparse initial task specification remains useful as the visual context evolves.

Recent hierarchical VLA systems also use spatial representations or planning intermediates for manipulation~\citep{li2025hamster, wu2025momanipvla, team2025gemini, zawalski2025robotic, Yang_2025_CVPR, wang2026vpvla, wei2026libra}. For example, HAMSTER predicts coarse 2D image-plane paths from RGB observations and language instructions, which then guide a separate low-level 3D-aware controller~\citep{li2025hamster}. In SP-VTP, a static first-frame object--target spatial prompt is given directly as the task specification, while the model predicts future relative 3D EE trajectory chunks from evolving egocentric observations. Language is omitted in this benchmark to isolate the role of spatial task specification.
%We do not claim spatial conditioning itself as new. SP-VTP instead isolates a specific combination: a static first-frame object--target prompt, evolving egocentric observations, and prediction of future relative 3D EE trajectory chunks.
%We omit language in this benchmark to isolate the contribution of spatial task specification, not to argue that spatial prompts should replace it.
SP-VTP brings together several challenges. The task specification is static, whereas the camera moves, the end effector occludes the scene, and the object relocates after grasping. Clutter and same-category distractors make the initial specification consequential, and the same object--target pair requires different motions across execution phases. SP-VTP evaluates whether the initial task specification remains useful as the egocentric view changes throughout execution.

%The task therefore evaluates persistent \emph{conditioning} under evolving views, not explicit persistent tracking: EgoSPT does not provide per-frame identity labels, and SPOT does not predict object correspondence. We restrict our claims accordingly to trajectory prediction rather than identity preservation or closed-loop manipulation success.

% \paragraph{EgoSPT: a dataset for spatially prompted manipulation.}
To study this problem, we introduce \emph{EgoSPT}, an egocentric dataset of spatially prompted manipulation trajectories collected with a modified Universal Manipulation Interface~(UMI)~\cite{chi2024universal}. EgoSPT provides first-frame object and target annotations, egocentric visual observations, recovered 3D EE trajectories, and a scene-stratified task-level split. The dataset focuses on a single fork pick-and-place primitive with five visually similar objects and nine receptacle targets, allowing us to study object–target specification across variations in layout and clutter. Accordingly, our evaluation focuses on spatial variation within a fixed task family rather than category- or skill-level generalization.

%It is deliberately controlled: all episodes instantiate one fork pick-and-place primitive with five visually similar objects and nine receptacle targets. This design isolates object--target specification under layout and clutter variation, but does not test category-level or skill-level generalization.
Building on EgoSPT, we introduce \emph{SPOT}, a spatially prompted object--target trajectory predictor. First-frame prompts specify the task, current observations and motion history provide execution context, and a decoder-style flow-matching head predicts a coherent future EE trajectory chunk. SPOT represents bounding boxes both as markings rendered on the first frame and as coordinate prompt tokens. These tokens attend to first-frame visual features to form a fixed task representation that conditions predictions together with the evolving observation context; SPOT does not contain an explicit object tracker.
Because egocentric backgrounds, camera motion, and object layouts are strongly correlated, sample-level random splits can overestimate performance. We therefore keep all episodes from a selected scene--task group within one partition while retaining all three scenes in both training and validation. Per-scene results characterize layout complexity rather than cross-scene generalization. We report continuous translation, rotation, and gripper errors together with thresholded endpoint criteria; all are open-loop prediction metrics rather than robot task success.
Our contributions are fourfold:
\begin{itemize}
  \item We define \emph{SP-VTP}, an open-loop setting for 3D visual trajectory prediction, in which object--target prompts specified in the first frame condition the prediction of future relative 3D EE trajectory chunks from evolving egocentric observations.

  \item We introduce \emph{EgoSPT}, a controlled egocentric dataset for spatially prompted trajectory prediction that provides object--target annotations, reconstructed 3D EE trajectories, and a comprehensive evaluation protocol for predicted trajectories.

  \item We propose \emph{SPOT}, a trajectory-prediction baseline that integrates visual and coordinate prompts with current observations and trajectory history, without relying on an explicit object tracker.

  \item We provide multi-seed statistics, comparisons with external VLA baselines, various prompting paradigms, and controlled prompt and execution-phase analyses.
\end{itemize}

\section{Related Work}

% \begin{NoHyper}

\paragraph{Goal-conditioned Robot Policies for Manipulation.}
Goal-conditioned manipulation policies aim to generate robot actions from sensor observation, under task specifications like task identifiers, goal images, or language instructions. 
In multi-task manipulation, task identifiers provide a compact way to indicate which task a policy should execute~\citep{yu2020meta, yang2020multitask}, but they require task intents to be discretized into fixed labels and lack scene-specific object-target grounding.
Goal images specify the desired final visual state~\citep{nair2018visual, srinivas2018universal, reuss2023goal, luo2025grounding}, but require access to an example of the completed scene and may entangle task intent with irrelevant visual details such as background, or object layout.

Language-conditioned policies provide more flexible interfaces for general robot manipulation~\citep{lynch2021language, shridhar2022cliport, shridhar2023perceiver, huang2023voxposer, goyal2024rvt}. 
More recently, generalist robot policies, particularly vision-language-action (VLA) models, have demonstrated the scalability of language-conditioned robot control across diverse tasks and embodiments~\citep{zitkovich2023rt, kim2024openvla, octo_2024, black2025pi}. 
However, language descriptions can remain ambiguous in cluttered egocentric scenes with multiple visually similar objects and candidate targets. Spatial representations have also been explored for conditioning manipulation policies. KITE grounds language instructions into 2D keypoints for downstream manipulation, while RT-Trajectory conditions visuomotor policies on coarse trajectory sketches~\citep{pmlr-v229-sundaresan23a, ICLR2024_c383e44d}. More recently, VP-VLA uses structured visual prompts as an interface between high-level planning and low-level control~\citep{wang2026vpvla}. SP-VTP studies a complementary setting in which a static first-frame object--target point or box directly specifies the task, while the model predicts future relative 3D EE trajectory chunks from evolving egocentric observations.

\begin{figure*}[t]
    \centering
    \includegraphics[width=0.97\linewidth]{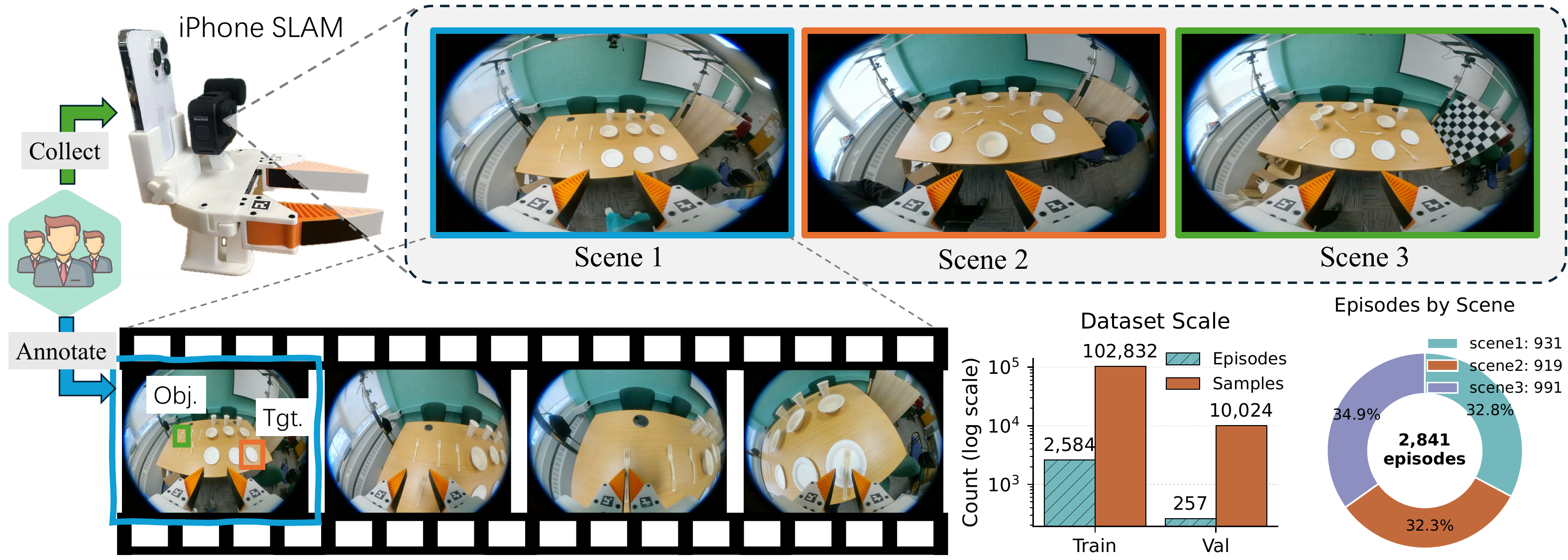}
    \caption{\textbf{Illustration of the EgoSPT dataset.} We use a modified UMI device equipped with an iPhone and a GoPro to collect EgoSPT, an egocentric visual trajectory dataset containing five forks and nine targets, including three plates, three bowls, and three cups. EgoSPT covers three scenes designed to evaluate different policy capabilities.}
    \label{fig:egospt-dataset}
\end{figure*}

\paragraph{Egocentric Manipulation Datasets.}
Egocentric manipulation datasets provide visual observations from the viewpoint of the acting agent, that are closely aligned with manipulation actions. Egocentric visual data has been studied in two related but distinct settings. Human-centered egocentric video datasets capture first-person observations from wearable cameras and support the study of human activities~\citep{grauman2022ego4d, damen2022rescaling, grauman2024ego}, 3D hand-object interactions~\citep{kwon2021h2o, tang2023egotracks, banerjee2025hot3d}, and imitation learning from human videos~\citep{kareer2025egomimic}. Manipulator-centered egocentric demonstration datasets, in contrast, place the camera on or near the manipulation interface, such as a gripper or robotic hand, so the visual stream is more directly aligned with manipulation execution and the action trajectories used for policy learning. Universal Manipulation Interface (UMI) is a representative setup in this direction, introducing a portable hand-held interface for collecting in-the-wild manipulation demonstrations and learning deployable visuomotor policies \citep{chi2024universal}. Subsequent UMI-style efforts further demonstrate the effectiveness of this action-aligned data-collection paradigm for training manipulation policies in multiple practical settings \citep{ha2024umionlegs, lin2025data, gupta2025umi,zhaxizhuoma2025fastumi, liu2025fastumi, seo2025legato, tao2025dexwild, liu2025maniwav, xu2025dexumi, nai2026humanoid}. Building on a modified UMI setup, EgoSPT collects egocentric manipulation videos with recovered EE trajectories and pairs them with first-frame object and target grounding annotations, turning demonstrations into data for spatially prompted visual trajectory prediction.

\paragraph{Egocentric Visual Trajectory Prediction.}
Egocentric visual trajectory prediction aims to infer future interaction motion from egocentric visual observations. Prior human-centered egocentric forecasting work studies image-space hand-object interaction prediction, where models forecast future hand motion and contact regions on active objects~\citep{liu2022joint, hatano2024emag, ma2025diff}. Other work extends egocentric forecasting to 3D, predicting action targets in 3D workspaces or future 3D hand trajectories from RGB videos~\citep{li2022egocentric, bao2023uncertainty, fang2024egopat3dv2}. These methods show that egocentric visual streams contain useful cues for future interaction, but they mainly forecast human-centered motion from observed video context.

Manipulator-centered imitation learning introduces a related but different trajectory prediction setting: future EE trajectory chunks, including pose and gripper state, serve as the action representation for manipulation policies~\citep{zhao2023learning, chi2025diffusion}. UMI-style systems use such trajectory chunks to learn deployable visuomotor policies from egocentric demonstrations~\citep{chi2024universal, ha2024umionlegs, nai2026humanoid}, where the task specification may be implicit in the demonstration or provided as a separate conditioning signal. SP-VTP focuses specifically on spatial task conditioning: a static first-frame object--target prompt specifies what to manipulate and where to place it, while the model predicts future relative EE trajectory chunks from subsequent egocentric observations.

\section{EgoSPT}

EgoSPT evaluates whether first-frame spatial prompts can be translated into future egocentric trajectories. As shown in Fig.~\ref{fig:egospt-dataset}, it contains 2,841 pick-and-place videos with egocentric observations and recovered end-effector (EE) trajectories, collected using a modified UMI device~\cite{chi2024universal,gupta2025umi}. The device integrates a GoPro with a $170^\circ$ fisheye lens for egocentric video capture and an iPhone for 6-DoF EE trajectory tracking. Leveraging the iPhone's visual--inertial SLAM system, the modified UMI provides more stable trajectory recovery than the original one.
Nine trained experts collect and annotate EgoSPT using this device. Each episode requires picking one of five visually similar forks and placing it into one of nine targets, consisting of three cups, three bowls, and three plates.

EgoSPT is organized into three scenes with increasing difficulty. Scene 1 contains structured layouts with 45 object--target combinations and 20 episodes per combination, testing object--target association under clean conditions. Scene 2 uses a cluttered layout with the same 45 combinations and 20 episodes per combination, evaluating robustness to distractors and spatial ambiguity. Scene 3 contains 22 randomly cluttered subscenes, each covering the 45 combinations with one episode per combination, enabling evaluation under diverse and low-data conditions. Together, these scenes form a progressive protocol for measuring behavior across structured and cluttered layouts. Because all three scenes occur in both training and validation, per-scene results are not used to claim cross-scene generalization.

% The dataset contains three scenes designed to evaluate different model capabilities. Scene 1 uses structured layouts and includes 45 object--target pick-and-place combinations, with 20 episodes collected for each combination. Scene 2 contains a cluttered layout with the same 45 object--target combinations and 20 episodes per combination, allowing us to evaluate robustness under object occlusion and visual distraction. Scene 3 consists of 22 randomly cluttered subscenes, each covering the 45 object--target combinations with only one episode per combination. This setting enables evaluation under more diverse and low-data conditions. Together, these scenes provide a progressive evaluation protocol, ranging from controlled structured environments to cluttered and highly diverse scenarios. Scene 1 mainly tests whether the model can learn reliable object--target associations under clean layouts. Scene 2 further examines generalization to cluttered environments where distractors and spatial ambiguity are present. Scene 3 evaluates whether the model can maintain performance when the data distribution becomes more diverse and the number of demonstrations per configuration is significantly reduced. This design allows us to systematically analyze both in-distribution performance and generalization ability across increasingly challenging manipulation settings.

Each video is approximately five seconds long at 30 fps and is downsampled to 10 fps, yielding about 50 trajectory steps per episode. 
Sliding-window training with horizon $H=16$ produces roughly 110K trajectory prediction samples. Each episode provides a first frame, 
current frames, object/target spatial prompts, trajectory history, and future relative trajectory chunks. More statistic results can be found in the Appendix.

The dataset stresses several realistic factors: multiple same-category object instances, three target categories, fisheye egocentric observations, handheld camera motion, and clutter variation. These factors make it a testbed for spatial task conditioning, not merely trajectory regression.

\section{SPOT}

\begin{figure*}[t]
    \centering
    \includegraphics[width=0.92\linewidth]{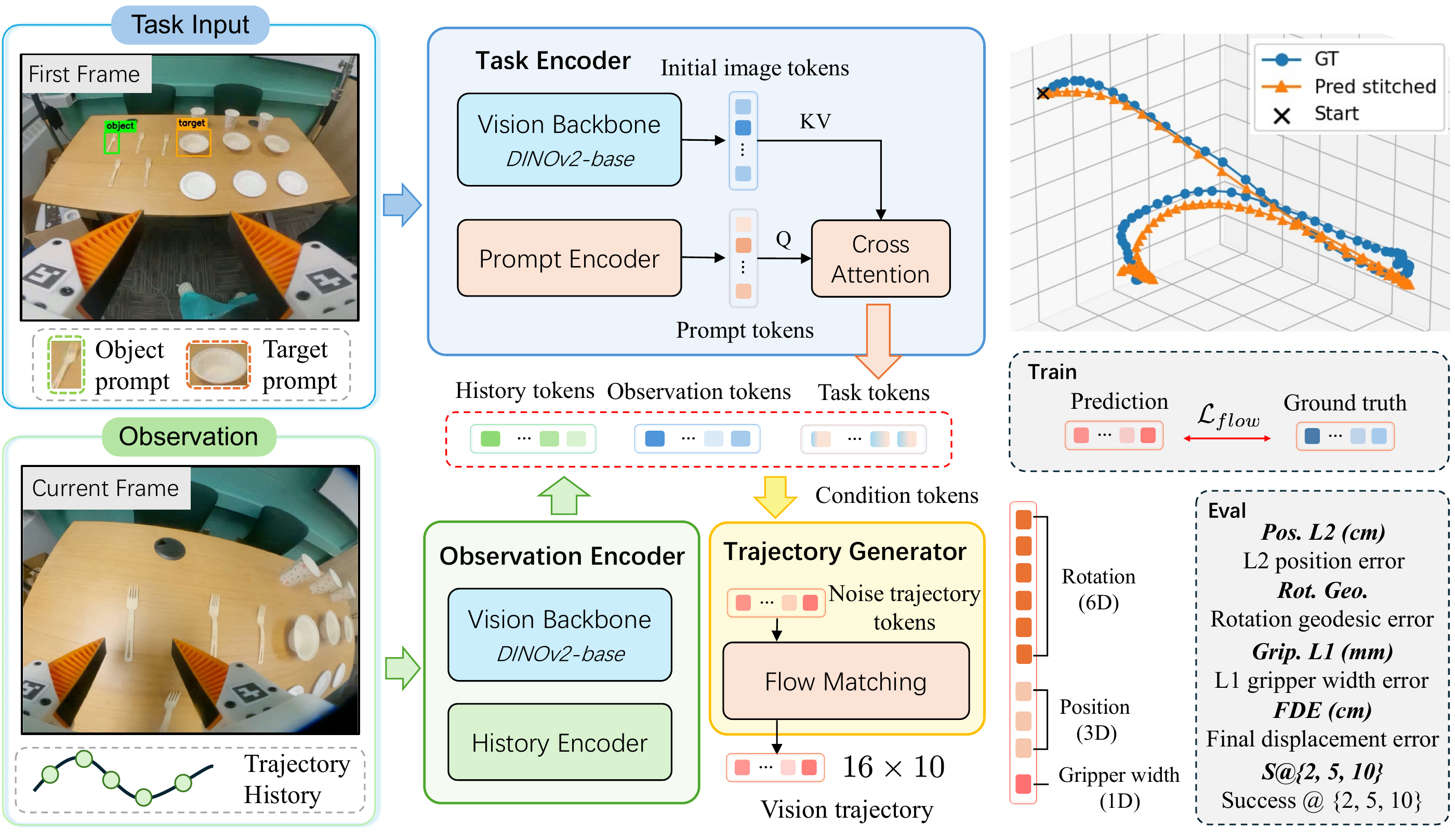}
    \caption{Overview of the proposed SPOT framework. Given a first-frame task specification, in which object and target boxes are rendered on the image and encoded as coordinate-prompt tokens, the task encoder extracts visual and prompt tokens and fuses them via cross-attention into a spatially prompted task representation. At each timestep, the observation encoder processes the current egocentric frame and action history, while the trajectory generator predicts future EE pose chunks using a flow-matching objective conditioned on the task and observation tokens. During training, the predicted trajectories are supervised by ground-truth trajectories using $\mathcal{L}_{\mathrm{flow}}$. During evaluation, performance is measured using Pos.~L2, Rot. Geo., Grip.~L1, and FDE.}
    \label{fig:spot_framework}
\end{figure*}

% \subsection{SPOT Overview}
As shown in Fig. \ref{fig:spot_framework}, SPOT is a spatial prompt conditioned trajectory policy composed of three modules: a task encoder, an observation encoder, and a trajectory generator. 
The task input combines visual prompts, obtained by rendering object and target boxes on the first frame, with coordinate prompts, which encode the same boxes as spatial tokens. The observation encoder represents the current egocentric frame and recent trajectory history. Conditioned on the resulting task and observation tokens, the trajectory generator predicts a future EE trajectory chunk.

% This design preserves the core principle of SPOT: spatial prompts define the task, current observations provide the execution context, and the trajectory decoder generates a coherent future trajectory chunk.

{\textbf{Problem Formulation.}}
SP-VTP takes a first-frame egocentric image $I_0$ with an object prompt $p_{\mathrm{obj}}$ and a target prompt $p_{\mathrm{tgt}}$, where each prompt is either a point or a box in normalized image coordinates. At timestep $t$, the policy observes the current egocentric frame $I_t$ and a raw history $H_t\in\mathbb{R}^{K\times10}$ of the previous $K$ actions. The goal is to predict a future trajectory chunk
$A_t\in\mathbb{R}^{H\times10} = \{a_t,\ldots,a_{t+H-1}\}, H=16$.
Each waypoint $a_{t+h}\in\mathbb{R}^{10}$ is parameterized as a 10D vector containing relative translation, 6D rotation, and gripper width. Let $T_t\in SE(3)$ denote the EE pose at timestep $t$ as a homogeneous transformation matrix in the world frame:
$T_t\in\mathbb{R}^{4\times4} =
\begin{bmatrix}
R_t & \mathbf{p}_t \\
\mathbf{0}^{\top} & 1
\end{bmatrix}
$,
where $R_t\in SO(3)$ is the EE orientation and $\mathbf{p}_t\in\mathbb{R}^3$ is its position. The relative pose target is computed in the current EE coordinate frame:
\begin{equation}
\Delta T_{t+h} = T_t^{-1}T_{t+h}
=
\begin{bmatrix}
\Delta R_{t+h} & \Delta \mathbf{p}_{t+h} \\
\mathbf{0}^{\top} & 1
\end{bmatrix},
\quad h \in [1,H].    
\end{equation}
Here, $T_t^{-1}$ transforms future poses from the world frame into the coordinate frame attached to the current EE, so $\Delta T_{t+h}$ represents the rigid motion needed to move from the current pose to the future pose. We then convert this relative transform into the 10D waypoint target:
\[
a_{t+h} =
\left[
\Delta \mathbf{p}_{t+h},
\mathrm{rot6d}(\Delta R_{t+h}),
g_{t+h}
\right]
\in \mathbb{R}^{10},
\]
where $\Delta \mathbf{p}_{t+h}\in\mathbb{R}^{3}$ is the relative translation, $\mathrm{rot6d}(\Delta R_{t+h})\in\mathbb{R}^{6}$ is the 6D representation of the relative rotation, and $g_{t+h}\in\mathbb{R}$ is the gripper width. This representation asks the model to predict where the EE should move next relative to its current state, rather than where it is in a global frame.

{\textbf{Task Encoder.}}
The task encoder processes the first frame $I_0$ together with the object and target prompts $(p_{\mathrm{obj}},p_{\mathrm{tgt}})$. In the default setting, $p_{\mathrm{obj}}$ and $p_{\mathrm{tgt}}$ are bounding boxes. We render these boxes on the first frame to obtain a visually prompted image $\tilde{I}_0$, and also keep their normalized box coordinates as coordinate prompts. A frozen DINOv2 ViT-B/14 \cite{oquab2024dinov2} backbone extracts and projects the first-frame features to the policy dimension $D=768$, yielding $F_0=[z_{\mathrm{img}};F_0^{\mathrm{patch}}]\in\mathbb{R}^{(N+1)\times D}$, where $F_0^{\mathrm{patch}}$ contains the $N$ patch tokens and $z_{\mathrm{img}}$ is the image-summary token, typically the projected DINOv2 CLS token.

Coordinate prompts are encoded geometrically. Each 2D prompt coordinate is mapped through Fourier positional features \cite{mildenhall2021nerf,tancik2020fourier} and an MLP (prompt encoder), and a learnable role embedding identifies whether the token belongs to the object or the target. A box prompt contributes two corner tokens for the object and two corner tokens for the target, while a point prompt contributes one object token and one target token. Let $M$ denote the total number of coordinate-prompt tokens, so $M=4$ for boxes and $M=2$ for points. The encoded prompts $Z_{\mathrm{prompt}}\in\mathbb{R}^{M\times D}$ query the first-frame visual tokens through a Transformer decoder \cite{vaswani2017attention}:
\begin{equation}
\begin{aligned}
    Z_{\mathrm{fused}} &= \mathrm{CrossAtt}(Q=Z_{\mathrm{prompt}},KV=F_0)
    \in\mathbb{R}^{M\times D},\\
    Z_{\mathrm{task}} &= [z_{\mathrm{img}};Z_{\mathrm{fused}}]
    \in\mathbb{R}^{(M+1)\times D}.
\end{aligned}
\end{equation}
This prompt-to-image cross-attention lets the object and target prompts actively read task-relevant visual information from the first frame. Appending the image-summary token to the $M$ fused prompt tokens gives a task representation containing both global scene context and grounded object/target evidence. In the default bounding-box setting, $M=4$ and $Z_{\mathrm{task}}\in\mathbb{R}^{5\times D}$.

{\textbf{Observation Encoder.}}
The observation encoder represents the execution state at timestep $t$. The current frame $I_t$ is encoded by the same frozen visual backbone and projection layer used by the task encoder, ensuring that first-frame task tokens and current-frame observation tokens live in a shared visual space. This shared visual encoding makes it easier for the trajectory generator to relate the original task specification to the current egocentric view.

To provide short-term motion context, a lightweight MLP, $\mathrm{Linear}(10,D)$--GELU--$\mathrm{Linear}(D,D)$, maps each raw action in $H_t$ to dimension $D$; learnable temporal-position and history-type embeddings are then added. Denoting this history encoder by $E_{\mathrm{hist}}$ and the current-frame image tokens by $F_t\in\mathbb{R}^{(N+1)\times D}$, we obtain
\[
\begin{aligned}
Z_{\mathrm{hist}} &= E_{\mathrm{hist}}(H_t)\in\mathbb{R}^{K\times D},\\
Z_{\mathrm{obs}} &= [F_t;Z_{\mathrm{hist}}]\in\mathbb{R}^{(N+1+K)\times D}.
\end{aligned}
\]
We set $K=4$ by default, giving $Z_{\mathrm{obs}}\in\mathbb{R}^{(N+5)\times D}$. If no trajectory history is used, the observation encoder returns only current-frame visual tokens. Otherwise, $Z_{\mathrm{obs}}$ jointly describes what the robot currently sees and how the EE has recently moved.

{\textbf{Trajectory Generator.}}
The trajectory generator predicts future trajectory chunks from the encoded task and execution context. It conditions on the concatenated task and observation tokens,
$Z_{\mathrm{cond}} = [Z_{\mathrm{task}}; Z_{\mathrm{obs}}]$.
In the default visual-prompt plus box-coordinate setting, $Z_{\mathrm{task}}$ consists of one image-summary token from the prompted first frame and four fused coordinate-prompt tokens, for five task tokens in total. For a $224\times224$ ViT-B/14 input, the current observation provides $N+1=257$ image tokens, while the history horizon contributes four trajectory-history tokens. Thus, $Z_{\mathrm{cond}}\in\mathbb{R}^{(N+10)\times D}$ contains 266 tokens in total, capturing the task specification, current egocentric context, and recent motion history.
% The trajectory generator predicts the future trajectory chunk from the encoded task and execution context. It is conditioned on the concatenation of task and observation tokens:
% $Z_{\mathrm{cond}} = [Z_{\mathrm{task}}; Z_{\mathrm{obs}}]$.
% Under the default visual-prompt plus box-coordinate setting, $Z_{\mathrm{task}}$ contains one image summary token from the visually prompted first frame and four fused coordinate prompt tokens. With a $224\times224$ ViT-B/14 input, the current observation contributes roughly 257 image tokens, and the default history horizon contributes four trajectory-history tokens. This produces a compact but expressive condition sequence that contains first-frame task specification, current visual context, and recent motion context.

SPOT supports both diffusion \cite{ho2020denoising, chi2025diffusion} and flow-matching \cite{lipmanflow}trajectory heads, using the same transformer decoder-style architecture in both cases. Noisy or interpolated trajectory tokens serve as the decoder targets, and $Z_{\mathrm{cond}}$ serves as the cross-attention memory. Each future waypoint can self-attend to other predicted waypoints and cross-attend to the task, observation, and history tokens.

The default head uses flow matching. Let $x_0$ denote the normalized ground-truth trajectory chunk and $\epsilon\sim\mathcal{N}(0,I)$ denote Gaussian noise. We sample $t\in[0,1]$ and construct a straight interpolation path
\begin{equation}
x_t = (1-t)x_0 + t\epsilon. 
\end{equation}
The target velocity is
\begin{equation}
v^\star(x_t,t)=\epsilon-x_0.    
\end{equation}
The flow head predicts this velocity from interpolated trajectory tokens and the condition memory:
\begin{equation}
\mathcal{L}_{\mathrm{flow}} =
\mathbb{E}\left[\|v_\theta(x_t,t,Z_{\mathrm{cond}})-(\epsilon-x_0)\|_2^2\right].    
\end{equation}
At inference time, SPOT starts from Gaussian noise and integrates the learned velocity field from noise to a trajectory chunk using a small number of Euler steps.
Future trajectory chunks are normalized by dataset-level waypoint mean and standard deviation before training. The trajectory head predicts in this normalized space, and predictions are denormalized after sampling. This balances translation, rotation, and gripper dimensions during optimization. During inference, the generated $H\times10$ trajectory chunk is denormalized and interpreted as relative EE motion.

\section{Experiments}
\begin{table*}[t]
  \centering
  \caption{Three-seed external-VLA comparison without language instance IDs. Four VLA baselines receive current observation image, $K=4$ history action chunks, and category language prompts without spatial prompts. We report mean $\pm$ sample standard deviation over seeds 42/0/1 on 10,024 samples, followed by scene-wise Pos.~L2/FDE (cm) and S@10 (\%).}
  \label{tab:external-baselines}
  \scriptsize
  \setlength{\tabcolsep}{3.5pt}
  \renewcommand{\arraystretch}{0.9}
  \resizebox{0.98\textwidth}{!}{%
    \begin{tabular}{lccccccc}
      \toprule
      Method & Pos. L2 (cm) & FDE (cm) & Rot. geo. ($^\circ$) & Grip. L1 (mm) & S@2 (\%) & S@5 (\%) & S@10 (\%) \\
      \midrule
      $\pi$0.5~\cite{pmlr-v305-black25a} & $14.25\!\pm\!0.18$ & $23.68\!\pm\!0.28$ & $15.53\!\pm\!0.02$ & $17.18\!\pm\!0.18$ & $0.33\!\pm\!0.02$ & $4.64\!\pm\!0.09$ & $20.67\!\pm\!0.47$ \\
     SmolVLA~\cite{shukor2025smolvla} & $9.36\!\pm\!0.22$ & $16.85\!\pm\!0.47$ & $11.06\!\pm\!0.16$ & $11.75\!\pm\!0.31$ & $1.53\!\pm\!0.17$ & $12.70\!\pm\!0.90$ & $37.21\!\pm\!1.72$ \\
      OpenVLA-OFT~\cite{kim2025fine} & $8.16\!\pm\!0.06$ & $15.33\!\pm\!0.14$ & $\mathbf{8.86\!\pm\!0.05}$ & $\mathbf{8.00\!\pm\!0.15}$ & $3.24\!\pm\!0.54$ & $17.14\!\pm\!0.68$ & $41.43\!\pm\!0.74$ \\
      GR00T N1.6~\cite{nvidia2025gr00tn16} & $10.48\!\pm\!1.07$ & $17.39\!\pm\!1.18$ & $11.60\!\pm\!0.47$ & $11.69\!\pm\!1.65$ & $0.88\!\pm\!0.29$ & $9.42\!\pm\!2.64$ & $32.42\!\pm\!4.47$ \\
      SPOT & $\mathbf{6.90\!\pm\!0.10}$ & $\mathbf{11.14\!\pm\!0.29}$ & $9.86\!\pm\!0.09$ & $12.76\!\pm\!1.25$ & $\mathbf{3.80\!\pm\!0.41}$ & $\mathbf{21.31\!\pm\!0.62}$ & $\mathbf{53.14\!\pm\!0.41}$ \\
      \bottomrule
    \end{tabular}%
  }

  \vspace{2pt}
  \setlength{\tabcolsep}{2.6pt}
  \resizebox{0.98\textwidth}{!}{%
    \begin{tabular}{lccccccccc}
      \toprule
      & \multicolumn{3}{c}{Scene 1 (4,333)} & \multicolumn{3}{c}{Scene 2 (3,079)} & \multicolumn{3}{c}{Scene 3 (2,612)} \\
      \cmidrule(lr){2-4}\cmidrule(lr){5-7}\cmidrule(lr){8-10}
      Method & Pos. & FDE & S@10 & Pos. & FDE & S@10 & Pos. & FDE & S@10 \\
      \midrule
      $\pi$0.5~\cite{pmlr-v305-black25a}
        & $11.63\!\pm\!0.25$ & $18.58\!\pm\!0.41$ & $25.49\!\pm\!0.93$
        & $14.79\!\pm\!0.21$ & $25.43\!\pm\!0.38$ & $22.59\!\pm\!0.40$
        & $17.95\!\pm\!0.17$ & $30.08\!\pm\!0.19$ & $10.40\!\pm\!0.28$ \\
      SmolVLA~\cite{shukor2025smolvla}
        & $7.35\!\pm\!0.24$ & $13.13\!\pm\!0.40$ & $48.17\!\pm\!1.31$
        & $10.22\!\pm\!0.42$ & $19.47\!\pm\!0.98$ & $36.05\!\pm\!3.57$
        & $11.69\!\pm\!0.07$ & $19.94\!\pm\!0.15$ & $20.41\!\pm\!0.57$ \\
      OpenVLA-OFT~\cite{kim2025fine}
        & $\mathbf{6.10\!\pm\!0.10}$ & $10.98\!\pm\!0.22$ & $53.55\!\pm\!1.89$
        & $8.80\!\pm\!0.10$ & $17.48\!\pm\!0.18$ & $38.75\!\pm\!0.20$
        & $10.83\!\pm\!0.19$ & $20.01\!\pm\!0.45$ & $24.50\!\pm\!1.21$ \\
      GR00T N1.6~\cite{nvidia2025gr00tn16}
        & $8.67\!\pm\!1.22$ & $13.26\!\pm\!1.46$ & $43.55\!\pm\!6.84$
        & $11.18\!\pm\!1.05$ & $19.71\!\pm\!1.23$ & $29.76\!\pm\!3.90$
        & $12.65\!\pm\!0.86$ & $21.51\!\pm\!0.92$ & $17.10\!\pm\!1.61$ \\
      SPOT
        & $6.71\!\pm\!0.26$ & $\mathbf{10.59\!\pm\!0.58}$ & $\mathbf{58.54\!\pm\!2.15}$
        & $\mathbf{6.25\!\pm\!0.07}$ & $\mathbf{10.88\!\pm\!0.28}$ & $\mathbf{53.53\!\pm\!1.40}$
        & $\mathbf{8.01\!\pm\!0.21}$ & $\mathbf{12.37\!\pm\!0.17}$ & $\mathbf{43.13\!\pm\!1.32}$ \\
      \bottomrule
    \end{tabular}%
  }
\end{table*}

\subsection{Experimental Setup}
\textbf{Protocol.}
All models train on Scenes 1--3 and share a held-out set of 257 episodes (10,024 action chunks); scene-wise results therefore measure layout complexity, not cross-scene generalization. Default SPOT uses rendered and coordinate object/target boxes, frozen DINOv2 ViT-B/14, cross-attention task fusion, $K=4$ history, and a flow-matching head. Prompt baselines use seeds 42/0/1 with split seed 42; diagnostic interventions use seed 42 and fixed initial action noise unless noted.

\textbf{Metrics.}
We report mean relative translation (Pos.~L2), final displacement (FDE), rotation geodesic (Rot.~geo.), and gripper-width L1 (Grip.~L1) errors in centimeters, degrees, and millimeters, respectively. Rot.~geo. averages the relative $\mathrm{SO}(3)$ angle after mapping 6D rotations via Gram--Schmidt. Endpoint metrics include Success@$X$cm ($X\in\{2,5,10\}$), gripper-state accuracy (G-end; 30 mm threshold), and Combined@5/10, which additionally requires rotation within $20^\circ$ and the correct gripper state. We emphasize that these metrics evaluate offline trajectory prediction rather than closed-loop task success.

\subsection{Comparison with External VLA Baselines}
We compare SPOT against SmolVLA~\cite{shukor2025smolvla}, OpenVLA-OFT~\cite{kim2025fine}, GR00T N1.6~\cite{nvidia2025gr00tn16}, and $\pi$0.5~\cite{pmlr-v305-black25a} using language-only prompts. These baselines receive current observation, an action history of $K=4$, and category-level language prompts, without access to the first frame or object and target boxes. All reported results are aggregated across three random seeds over 10,024 samples.

Table~\ref{tab:external-baselines} shows that SPOT has the best overall translation performance and the highest S@2/5/10 scores. Its advantage grows with layout complexity: compared to OpenVLA-OFT, the FDE margin increases from 0.39 cm in Scene 1 to 6.60 and 7.64 cm in Scenes 2 and 3, respectively, while the S@10 margin increases from 4.99$\%$ to 14.78$\%$ and 18.63$\%$.  This pattern suggests that \textit{instance grounding becomes a greater bottleneck in cluttered environments, rather than reflecting differences in cross-scene generalization}. OpenVLA-OFT retains better Rot.~geo. and Grip.~L1 performance, indicating that spatial prompting primarily improves \emph{where} to move rather than fine-grained pose control. However, differences in evaluation protocols preclude conclusions about end-to-end VLA training without instance identifiers. Results across three scenes also reveal a clear difficulty hierarchy: Scene 3 has the most complex layout, followed by Scenes 2 and 1.

\subsection{Prompting Baselines}
We compare five task specifications: \textsc{No Prompt}, which replaces object/target information with learned null tokens; \textsc{BBox Prompt}, which uses coordinate box tokens; \textsc{Visual Prompt}, which renders the boxes on the first frame; \textsc{Target Only}, which provides only the target’s visual prompt and bounding-box information; and \textsc{BBox+Visual}, which combines coordinate and rendered prompts.
\begin{table*}[t]
  \centering
  \caption{Three-seed prompting results, reported as mean $\pm$ sample standard deviation. Lower is better for the four error metrics; higher is better for endpoint translation success. Paired-bootstrap confidence intervals are reported in Appendix Table~\ref{tab:baseline-confidence-intervals}.}
  \label{tab:baseline-results}
  \small
  \setlength{\tabcolsep}{5pt}
  \resizebox{0.95\textwidth}{!}{%
    \begin{tabular}{lccccccc}
      \toprule
      Prompt & Pos. L2 (cm) & FDE (cm) & Rot. geo. ($^\circ$) & Grip. L1 (mm) & S@2 (\%) & S@5 (\%) & S@10 (\%) \\
      \midrule
      No prompt     & $9.39\!\pm\!0.26$ & $17.75\!\pm\!0.39$ & $10.27\!\pm\!0.16$ & $11.91\!\pm\!0.99$ & $4.31\!\pm\!0.63$ & $19.47\!\pm\!1.17$ & $39.19\!\pm\!0.97$ \\
      Target only   & $7.07\!\pm\!0.02$ & $11.57\!\pm\!0.12$ & $10.22\!\pm\!0.27$ & $13.09\!\pm\!0.92$ & $3.35\!\pm\!0.08$ & $20.16\!\pm\!0.58$ & $50.14\!\pm\!0.56$ \\
      Visual        & $6.83\!\pm\!0.18$ & $11.25\!\pm\!0.50$ & $9.49\!\pm\!0.10$ & $12.08\!\pm\!1.10$ & $3.99\!\pm\!0.34$ & $22.66\!\pm\!0.94$ & $52.93\!\pm\!3.17$ \\
      BBox          & $7.15\!\pm\!0.07$ & $11.85\!\pm\!0.14$ & $10.33\!\pm\!0.14$ & $13.05\!\pm\!1.21$ & $3.00\!\pm\!0.17$ & $19.01\!\pm\!0.19$ & $49.02\!\pm\!0.36$ \\
      BBox + visual & $6.90\!\pm\!0.10$ & $11.14\!\pm\!0.29$ & $9.86\!\pm\!0.09$ & $12.76\!\pm\!1.25$ & $3.80\!\pm\!0.41$ & $21.31\!\pm\!0.62$ & $53.14\!\pm\!0.41$ \\
      \bottomrule
    \end{tabular}%
  }
\end{table*}

Table~\ref{tab:baseline-results} shows that grounding primarily reduces large endpoint errors. Compared with No Prompt, BBox+Visual lowers Pos.~L2 and FDE by 2.49 cm and 6.61 cm, respectively, and improves S@10 by 13.95$\%$, while leaving S@2 largely unchanged. This suggests that spatial grounding helps resolve ambiguous object--target associations but does not substantially refine predictions that are already close to the target. Visual Prompt achieves the best performance on several mean metrics, whereas BBox alone is generally weaker, indicating that image-space cues align more naturally with the frozen visual backbone. Coordinate tokens nevertheless improve prediction stability: BBox+Visual achieves the highest S@10 and reduces the across-seed standard deviation from 3.17$\%$ to 0.41$\%$ relative to Visual Prompt. Target only captures most of the FDE improvement, while the full prompt further improves Pos.~L2 and S@5/10, suggesting that the target specifies the destination and the object context helps disambiguate the trajectory path. Appendix Table~\ref{tab:baseline-confidence-intervals} reports significance tests.

\subsection{Spatial Prompt Dependence and Robustness}
To determine whether SPOT relies on spatial prompts rather than dataset priors, we perturb the spatial prompts at inference time while keeping the observation, motion history, ground truth, model checkpoint, and initial action noise fixed. For the shuffled and role-swapped conditions, we redraw the rendered visual prompt on the source-resolution first frame and update the corresponding coordinate tokens accordingly.

\par\medskip
\noindent\begin{minipage}{\columnwidth}
  \centering
  \captionof{table}{Inference-time prompt interventions on the same seed BBox+Visual checkpoint. Observation, history, ground truth, and initial action noise are fixed.  G-end denotes endpoint gripper-state accuracy after binarizing gripper width at 30mm. Combined@10 denotes the percentage of action chunks whose endpoint simultaneously satisfies FDE $\leq 10$ cm, Rot. geo $\leq 20^\circ$, and correct gripper state. Both are offline endpoint metrics rather than closed-loop task-success rates.}
  \label{tab:prompt-interventions}
  \scriptsize
  \setlength{\tabcolsep}{3pt}
  \resizebox{\columnwidth}{!}{%
    \begin{tabular}{lcccc}
      \toprule
      Prompt condition & FDE (cm) & Rot. geo. ($^\circ$) & G-end (\%) & Combined@10 (\%) \\
      \midrule
      Shuffled object       & 13.8424 & 10.90 & 80.56 & 29.93 \\
      Shuffled target       & 18.4667 & 11.09 & 79.34 & 26.68 \\
      Object/target swap    & 22.0824 & 11.23 & 79.62 & 24.77 \\
      Correct BBox + visual & \textbf{11.4655} & \textbf{9.95} & \textbf{81.04} & \textbf{38.67} \\
      \bottomrule
    \end{tabular}%
  }
\end{minipage}
\par\medskip

Table~\ref{tab:prompt-interventions} shows that shuffling the object/target raises FDE by 20.74\%/61.04\%, which indicates that target placement is substantially more challenging than object pickup. Swapping the object and target roles produces an even larger FDE increase of 92.60\%. This degradation exceeds that caused by either shuffle, even though both regions remain visible, indicating that SPOT relies on their semantic roles rather than mere visual saliency.  By contrast, Rot.~geo. changes by at most $1.28^\circ$, and G-end changes by only 1.70\%, suggesting that prompt corruption redirects the overall trajectory much more strongly than it affects local pose or gripper state. 
Additional robustness results under prompt scaling and jitter are provided in the Appendix.

\par\medskip
\noindent\begin{minipage}{\columnwidth}
  \centering
  \captionof{table}{Five-stage endpoint analysis. Each chunk is grouped by its endpoint phase; phase definitions and segmentation details are given in Appendix~\ref{sec:phase-segmentation}. Values are FDE (cm); lower is better.}
  \label{tab:phase-analysis}
  \scriptsize
  \setlength{\tabcolsep}{2pt}
  \resizebox{\columnwidth}{!}{%
    \begin{tabular}{lccccc}
      \toprule
      Condition & Approach & Pre-grasp & Grasp/occl. & Transport & Placement \\
      \midrule
      No prompt       & 12.41 & {9.42} & 7.05 & 15.63 & 25.41 \\
      Shuffled object & 18.64 & 14.30 & 10.25 & 13.13 & 15.42 \\
      Shuffled target & \textbf{11.54} & \textbf{9.28} & 7.55 & 16.69 & 27.21 \\
      Role swap       & 12.00 & 13.74 & 11.08 & 19.18 & 32.09 \\
      Target only     & 14.28 & 11.54 & 8.84 & 12.33 & \textbf{11.36} \\
      BBox + visual   & 12.16 & 9.79 & \textbf{6.97} & \textbf{12.29} & 12.83 \\
      \bottomrule
    \end{tabular}%
  }
\end{minipage}
\par\medskip

\subsection{Role-Specific Effects Across Execution}

We segment each validation episode using ground-truth gripper width and EE translation speed into Approach (initial motion), Pre-grasp (final deceleration), Grasp/Occlusion~(closing and initial lift), Transport (post-grasp transfer), and Placement (final target approach and release). Each trajectory chunk’s FDE is assigned to the phase containing its future endpoint timestep. The detection thresholds, fallback rules, and segmentation diagnostics are provided in Appendix~\ref{sec:phase-segmentation}.

Table~\ref{tab:phase-analysis} reveals when spatial ambiguity matters most. BBox+Visual performs similarly to No Prompt during the early phases but reduces FDE during Transport and Placement by 21.4\% and 49.5\%, respectively. This pattern suggests that the current observation and motion history largely constrain early motion, whereas spatial grounding becomes critical when selecting and approaching the destination. Shuffling the object prompt primarily degrades performance during the first three phases, while shuffling the target prompt increases Transport and Placement FDE by 35.8\% and 112.1\%, respectively. Target Only performs best during Placement, suggesting that stage-aware fusion could downweight object cues once the grasp has been completed. Overall, object cues primarily guide acquisition, whereas target cues guide delivery, rather than both contributing uniformly throughout the trajectory.

We further aggregate performance into Pick and Place stages using two complementary definitions: a task-half split and contact-centered windows that include chunk endpoints within $\pm15$ raw video frames (approximately $\pm0.5$~s) of the detected pick or place contact. Stage boundaries are jointly estimated from the EE position and gripper width.

\begin{table}[t]
  \centering
  \caption{Pick/place endpoint analysis under two phase definitions. FDE is in centimeters; S@2/5/10 are offline endpoint criteria reported in percent.}
  \label{tab:pick-place-analysis}
  \scriptsize
  \setlength{\tabcolsep}{2pt}
  \resizebox{\columnwidth}{!}{%
    \begin{tabular}{llcccc}
      \toprule
      Split & Condition & Pick FDE & Place FDE & Pick S@2/5/10 & Place S@2/5/10 \\
      \midrule
      \multirow{3}{*}{Task half}
        & No prompt     & 10.76 & 27.32 & \textbf{5.5}/\textbf{26.6}/56.0 & 2.5/7.7/14.7 \\
        & Target only   & 10.81 & \textbf{12.40} & 2.3/20.3/52.7 & \textbf{3.0/14.8/43.8} \\
        & BBox + visual & \textbf{9.59} & 13.86 & 3.5/23.6/\textbf{61.6} & 2.3/13.5/37.4 \\
      \midrule
      \multirow{3}{*}{Contact $\pm15$}
        & No prompt     & 8.32 & 24.79 & \textbf{9.4}/\textbf{39.1}/69.5 & 4.0/12.9/22.9 \\
        & Target only   & 9.78 & \textbf{11.50} & 3.4/26.2/58.7 & \textbf{3.9/20.1/49.8} \\
        & BBox + visual & \textbf{7.88} & 13.46 & 5.7/32.6/\textbf{74.3} & 2.7/18.2/40.9 \\
      \bottomrule
    \end{tabular}%
  }
\end{table}

Table~\ref{tab:pick-place-analysis} confirms this phase-specific pattern under both definitions. BBox+Visual achieves the lowest Pick FDE and the highest Pick S@10, whereas No Prompt retains the highest Pick S@2/5. Spatial prompting therefore removes gross pickup errors without improving the finest local accuracy, consistent with the aggregate success thresholds. During Place, BBox+Visual reduces the No Prompt FDE by nearly half under both splits, while Target Only achieves the best overall performance. The consistency across the task-half, contact-window, and high-confidence-boundary evaluations suggests that this role separation is unlikely to be an artifact of phase detection. Nevertheless, these results measure offline trajectory prediction rather than closed-loop success.

\subsection{Architecture Ablations}
\begin{figure}[t]
    \centering
    \includegraphics[width=\linewidth]{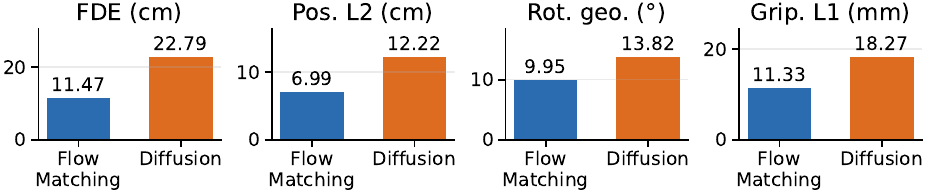}
    \caption{\textbf{Trajectory head ablation.} Flow matching consistently outperforms diffusion. Lower is better.}
    \label{fig:ablation-head-full}
\end{figure}

\textbf{Trajectory head.} Figure~\ref{fig:ablation-head-full} shows that the choice of trajectory head has the largest effect. Compared with diffusion, flow matching reduces FDE and Pos.~L2 by 49.7\% and 42.8\%, respectively, while lowering Rot.~geo. and Grip.~L1. Consistent endpoint, waypoint, rotation, and gripper gains rule out an endpoint-only effect.

\begin{figure}[t]
    \centering
    \includegraphics[width=\linewidth]{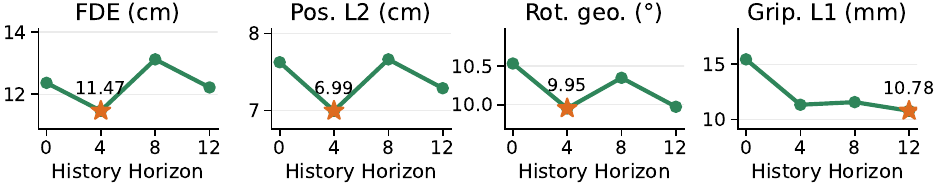}
    \caption{\textbf{History horizon ablation.} Stars mark the best value for each metric. Lower is better.}
    \label{fig:ablation-history-full}
\end{figure}

\textbf{History horizon.} Figure~\ref{fig:ablation-history-full} is distinctly non-monotonic. Moving from $K=0$ to 4 lowers FDE by 7.3\% and Grip.~L1 by 26.7\%, showing that recent motion is especially informative for gripper state. $K=8$ worsens all four metrics, while $K=12$ recovers rotation and gives the best Grip.~L1 but remains worse than $K=4$ in translation. Thus, older states may preserve phase information but dilute endpoint evidence; $K=4$ offers the best balance.

\begin{figure}[t]
    \centering
    \includegraphics[width=\linewidth]{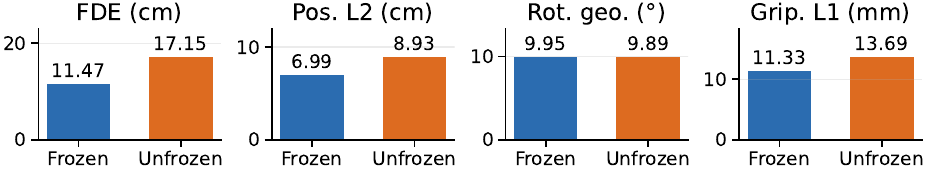}
    \caption{\textbf{Visual backbone tuning ablation.} The frozen DINOv2 encoder outperforms end-to-end tuning. Lower is better.}
    \label{fig:ablation-dinov2-base-tuning-full}
\end{figure}

\textbf{Backbone tuning.} In Figure~\ref{fig:ablation-dinov2-base-tuning-full}, freezing DINOv2 improves FDE, Pos.~L2, and Grip.~L1 by 33.1\%, 21.7\%, and 17.3\%; fine-tuning improves Rot.~geo. by only $0.07^\circ$. Preserving pretrained spatial features thus outweighs the negligible orientation gain on this dataset. Because the unfrozen model is prone to overfitting. 

Overall, these ablations favor frozen features, $K=4$ history, and flow generation; encoder-family/size analyses are presented in the Appendix.

\subsection{Qualitative Evaluation}
\begin{figure}[t]
    \centering
    \includegraphics[width=\linewidth]{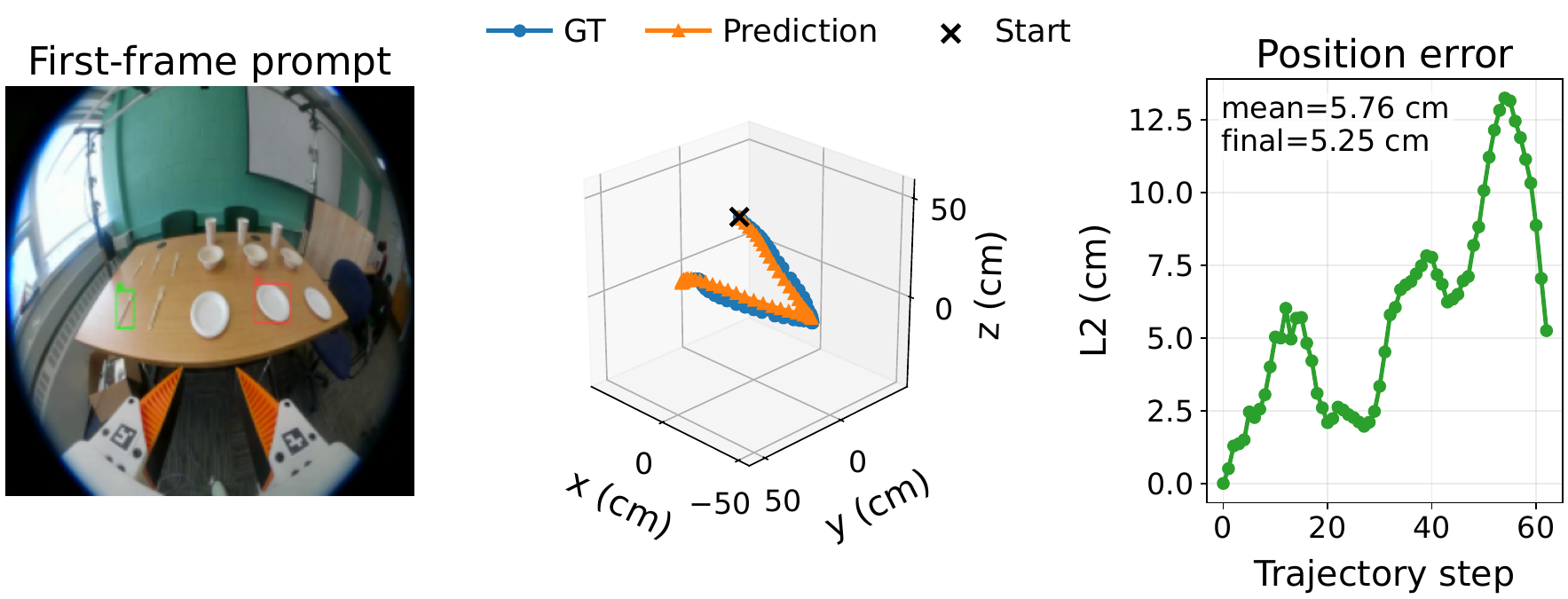}
    \caption{\textbf{Full trajectory visualization.} We show stitched predictions over full episodes together with ground-truth 3D trajectories, first-frame spatial prompts, and per-frame position errors.}
    \label{fig:full-trajectory-main}
\end{figure}

\begin{figure}[t]
    \centering
    \includegraphics[width=1\linewidth]{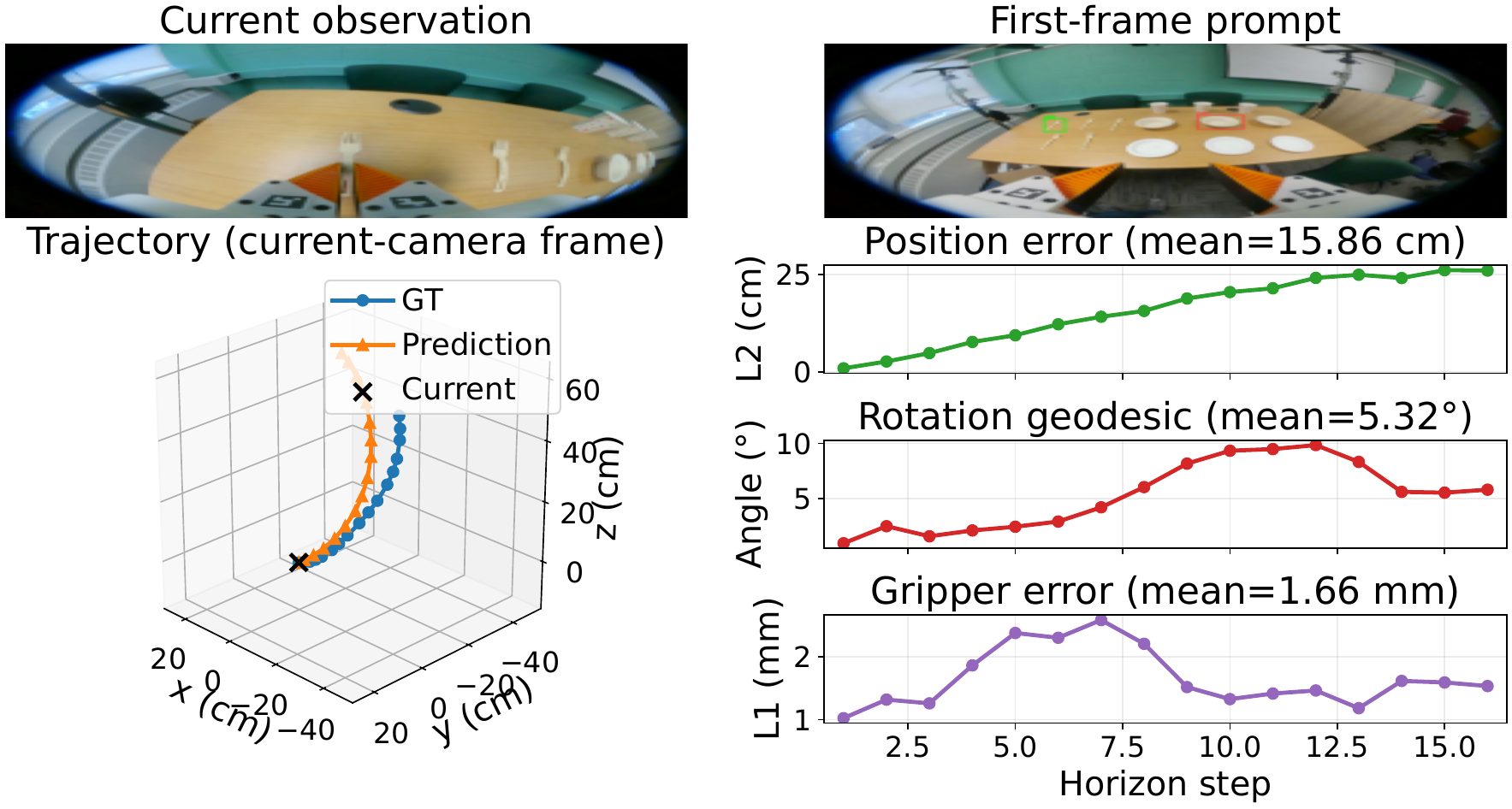}
    \caption{\textbf{First chunk visualization.} We show the first-frame prompt, current observation, predicted and ground-truth trajectory chunks, and chunk-level Pos. L2, Rot. L2, and Grip. L1 curves.}
    \label{fig:first-chunk-main}
\end{figure}

Figures~\ref{fig:full-trajectory-main} and \ref{fig:first-chunk-main} show that SPOT recovers coarse motion geometry but accumulates error across stitched chunks. Together with strong S@10 but limited S@2 gains, this separates task grounding from fine control: prompts select the motion mode, motivating closed-loop replanning for drift correction. These examples do not establish closed-loop success; more scenes appear in the Appendix.

\vspace{-5pt}
\section{Conclusion}
\vspace{-5pt}
We introduce \emph{Spatially Prompted Visual Trajectory Prediction} (SP-VTP), EgoSPT with spatial grounding and recovered 3D motion, and SPOT, which fuses visual and coordinate prompts with observation and motion history. Across three seeds, spatial prompts reduce translation and endpoint errors, with Visual and BBox+Visual prompts offering complementary advantages. Controlled interventions further localize object-cue effects to approach/grasp and target-cue effects to transport/placement. Spatial prompting is therefore an effective interface for visually grounded trajectory prediction; closed-loop execution remains future work.

% We introduced \emph{Spatially Prompted Visual Trajectory Prediction} (SP-VTP), where first-frame object and target prompts condition egocentric EE trajectory prediction. We instantiated this setting with EgoSPT and SPOT, a prompt-centric policy that fuses visual and coordinate prompts with current observation and motion history for flow-based generation. Scene-aware experiments show that spatial prompts improve trajectory accuracy, with combined visual and coordinate prompts performing best. These results establish spatial prompting as a compact and effective interface for visually grounded manipulation.

{
  \small
  \bibliographystyle{ieeenat_fullname}
  \bibliography{references}

@String(IJCV = {Int. J. Comput. Vis.})

@String(CVPR= {IEEE Conf. Comput. Vis. Pattern Recog.})

@String(ICCV= {Int. Conf. Comput. Vis.})

@String(ECCV= {Eur. Conf. Comput. Vis.})

@String(NIPS= {Adv. Neural Inform. Process. Syst.})

@String(ICLR = {Int. Conf. Learn. Represent.})

@String(ICML = {Int. Conf. Machi. Learn.})

@String(TMLR = {Trans. Mach. Learn. Res.})

@String(CORL = {Conf. Robot Learn.})

@String(RSS = {Robotics: Science and Systems})

@String(ICRA = {IEEE Int. Conf. Robot. Autom.})

@String(IROS = {IEEE/RSJ Int. Conf. Intell. Robots Syst.})

@String(RAL = {IEEE Robot. Autom. Lett.})

@String(IJRR = {Int. J. Robot. Res.})

@inproceedings{ICLR2024_c383e44d,
 author = {Gu, Jiayuan and Kirmani, Sean and Wohlhart, Paul and Lu, Yao and Gonzalez Arenas, Montserrat and Rao, Kanishka and Yu, Wenhao and Fu, Chuyuan and Gopalakrishnan, Keerthana and Xu, Zhuo and Sundaresan, Priya and Xu, Peng and Su, Hao and Hausman, Karol and Finn, Chelsea and Vuong, Quan and Xiao, Ted},
 booktitle = {International Conference on Learning Representations},
 editor = {B. Kim and Y. Yue and S. Chaudhuri and K. Fragkiadaki and M. Khan and Y. Sun},
 pages = {44918--44941},
 title = {RT-Trajectory: Robotic Task Generalization via Hindsight Trajectory Sketches},
 url = {https://proceedings.iclr.cc/paper_files/paper/2024/file/c383e44d9a878d1982d9abb838bd5d8a-Paper-Conference.pdf},
 volume = {2024},
 year = {2024}
}

@InProceedings{pmlr-v229-sundaresan23a,
  title = 	 {KITE: Keypoint-Conditioned Policies for Semantic Manipulation},
  author =       {Sundaresan, Priya and Belkhale, Suneel and Sadigh, Dorsa and Bohg, Jeannette},
  booktitle = 	 {Proceedings of The 7th Conference on Robot Learning},
  pages = 	 {1006--1021},
  year = 	 {2023},
  editor = 	 {Tan, Jie and Toussaint, Marc and Darvish, Kourosh},
  volume = 	 {229},
  series = 	 {Proceedings of Machine Learning Research},
  month = 	 {06--09 Nov},
  publisher =    {PMLR},
  url = 	 {https://proceedings.mlr.press/v229/sundaresan23a.html},
}

@misc{shukor2025smolvla,
      title={SmolVLA: A Vision-Language-Action Model for Affordable and Efficient Robotics}, 
      author={Mustafa Shukor and Dana Aubakirova and Francesco Capuano and Pepijn Kooijmans and Steven Palma and Adil Zouitine and Michel Aractingi and Caroline Pascal and Martino Russi and Andres Marafioti and Simon Alibert and Matthieu Cord and Thomas Wolf and Remi Cadene},
      year={2025},
      eprint={2506.01844},
      archivePrefix={arXiv},
      primaryClass={cs.LG},
      url={https://arxiv.org/abs/2506.01844}, 
}

@article{kim2025fine,
  title={Fine-Tuning Vision-Language-Action Models: Optimizing Speed and Success},
  author={Kim, Moo Jin and Finn, Chelsea and Liang, Percy},
  journal={arXiv preprint arXiv:2502.19645},
  year={2025}
}

@misc{nvidia2025gr00tn16,
    author       = {{NVIDIA GEAR Team}},
    title        = {{GR00T N1.6}: An Improved Open Foundation Model for Generalist Humanoid Robots},
    year         = {2025},
    howpublished = {NVIDIA Research},
    note         = {Released December 15, 2025},
    url = {https://research.nvidia.com/labs/gear/gr00t-n1_6/}
}

@InProceedings{pmlr-v305-black25a,
  title = 	 {$\pi_{0.5}$: a Vision-Language-Action Model with Open-World Generalization},
  author =       {Black, Kevin and Brown, Noah and Darpinian, James and Dhabalia, Karan and Driess, Danny and Esmail, Adnan and Equi, Michael Robert and Finn, Chelsea and Fusai, Niccolo and Galliker, Manuel Y. and Ghosh, Dibya and Groom, Lachy and Hausman, Karol and ichter, brian and Jakubczak, Szymon and Jones, Tim and Ke, Liyiming and LeBlanc, Devin and Levine, Sergey and Li-Bell, Adrian and Mothukuri, Mohith and Nair, Suraj and Pertsch, Karl and Ren, Allen Z. and Shi, Lucy Xiaoyang and Smith, Laura and Springenberg, Jost Tobias and Stachowicz, Kyle and Tanner, James and Vuong, Quan and Walke, Homer and Walling, Anna and Wang, Haohuan and Yu, Lili and Zhilinsky, Ury},
  booktitle = 	 {Proceedings of The 9th Conference on Robot Learning},
  pages = 	 {17--40},
  year = 	 {2025},
  editor = 	 {Lim, Joseph and Song, Shuran and Park, Hae-Won},
  volume = 	 {305},
  series = 	 {Proceedings of Machine Learning Research},
  month = 	 {27--30 Sep},
  publisher =    {PMLR},
  url = 	 {https://proceedings.mlr.press/v305/black25a.html}
}

@article{wang2026vpvla,
  title={VP-VLA: Visual Prompting as an Interface for Vision-Language-Action Models},
  author={Wang, Zixuan and Chen, Yuxin and Liu, Yuqi and Ye, Jinhui and Chen, Pengguang and Lu, Changsheng and Liu, Shu and Jia, Jiaya},
  journal={arXiv preprint arXiv:2603.22003},
  year={2026}
}

@InProceedings{Yang_2025_CVPR,
    author    = {Yang, Jianwei and Tan, Reuben and Wu, Qianhui and Zheng, Ruijie and Peng, Baolin and Liang, Yongyuan and Gu, Yu and Cai, Mu and Ye, Seonghyeon and Jang, Joel and Deng, Yuquan and Gao, Jianfeng},
    title     = {Magma: A Foundation Model for Multimodal AI Agents},
    booktitle = CVPR,
    month     = {June},
    year      = {2025},
    pages     = {14203-14214}
}

@inproceedings{wu2025momanipvla,
  title={Momanipvla: Transferring vision-language-action models for general mobile manipulation},
  author={Wu, Zhenyu and Zhou, Yuheng and Xu, Xiuwei and Wang, Ziwei and Yan, Haibin},
  booktitle=CVPR,
  pages={1714--1723},
  year={2025}
}

@article{wei2026libra,
  title={Libra-VLA: Achieving Learning Equilibrium via Asynchronous Coarse-to-Fine Dual-System},
  author={Wei, Yifei and Zhong, Linqing and Liu, Yi and Lu, Yuxiang and He, Xindong and Yao, Maoqing and Ren, Guanghui},
  journal={arXiv preprint arXiv:2604.24921},
  year={2026}
}

@inproceedings{zawalski2025robotic,
  title={Robotic Control via Embodied Chain-of-Thought Reasoning},
  author={Zawalski, Micha{\l} and Chen, William and Pertsch, Karl and Mees, Oier and Finn, Chelsea and Levine, Sergey},
  booktitle=CORL,
  pages={3157--3181},
  year={2025},
  organization={PMLR}
}

@article{chi2025diffusion,
  title={Diffusion policy: Visuomotor policy learning via action diffusion},
  author={Chi, Cheng and Xu, Zhenjia and Feng, Siyuan and Cousineau, Eric and Du, Yilun and Burchfiel, Benjamin and Tedrake, Russ and Song, Shuran},
  journal=IJRR,
  volume={44},
  number={10-11},
  pages={1684--1704},
  year={2025},
  publisher={Sage Publications Sage UK: London, England}
}

@inproceedings{zhao2023learning,
title={Learning Fine-Grained Bimanual Manipulation with Low-Cost Hardware},
author={Tony Z. Zhao and Vikash Kumar and Sergey Levine and Chelsea Finn},
booktitle=RSS,
year={2023}
}

@inproceedings{liu2022joint,
  title={Joint hand motion and interaction hotspots prediction from egocentric videos},
  author={Liu, Shaowei and Tripathi, Subarna and Majumdar, Somdeb and Wang, Xiaolong},
  booktitle=CVPR,
  pages={3282--3292},
  year={2022}
}

@inproceedings{hatano2024emag,
  title={Emag: Ego-motion aware and generalizable 2d hand forecasting from egocentric videos},
  author={Hatano, Masashi and Hachiuma, Ryo and Saito, Hideo},
  booktitle=ECCV,
  pages={119--136},
  year={2024}
}

@inproceedings{ma2025diff,
  title={Diff-ip2d: Diffusion-based hand-object interaction prediction on egocentric videos},
  author={Ma, Junyi and Chen, Xieyuanli and Xu, Jingyi and Wang, Hesheng},
  booktitle=IROS,
  pages={4291--4298},
  year={2025},
  organization={IEEE}
}

@inproceedings{li2022egocentric,
  title={Egocentric prediction of action target in 3d},
  author={Li, Yiming and Cao, Ziang and Liang, Andrew and Liang, Benjamin and Chen, Luoyao and Zhao, Hang and Feng, Chen},
  booktitle=CVPR,
  pages={20971--20980},
  year={2022},
  organization={IEEE}
}

@inproceedings{bao2023uncertainty,
  title={Uncertainty-aware state space transformer for egocentric 3d hand trajectory forecasting},
  author={Bao, Wentao and Chen, Lele and Zeng, Libing and Li, Zhong and Xu, Yi and Yuan, Junsong and Kong, Yu},
  booktitle=ICCV,
  pages={13702--13711},
  year={2023}
}

@inproceedings{fang2024egopat3dv2,
  title={Egopat3dv2: Predicting 3d action target from 2d egocentric vision for human-robot interaction},
  author={Fang, Irving and Chen, Yuzhong and Wang, Yifan and Zhang, Jianghan and Zhang, Qiushi and Xu, Jiali and He, Xibo and Gao, Weibo and Su, Hao and Li, Yiming and others},
  booktitle=ICRA,
  pages={3036--3043},
  year={2024},
  organization={IEEE}
}

@inproceedings{kareer2025egomimic,
  title={Egomimic: Scaling imitation learning via egocentric video},
  author={Kareer, Simar and Patel, Dhruv and Punamiya, Ryan and Mathur, Pranay and Cheng, Shuo and Wang, Chen and Hoffman, Judy and Xu, Danfei},
  booktitle=ICRA,
  pages={13226--13233},
  year={2025},
  organization={IEEE}
}

@article{li2025hamster,
  title={HAMSTER: Hierarchical Action Models For Open-World Robot Manipulation},
  author={Li, Yi and Deng, Yuquan and Zhang, Jesse and Jang, Joel and Memmel, Marius and Yu, Raymond and Garrett, Caelan Reed and Ramos, Fabio and Fox, Dieter and Li, Anqi and Gupta, Abhishek and Goyal, Ankit},
  journal={arXiv preprint arXiv:2502.05485},
  year={2025}
}

@article{tang2023egotracks,
  title={Egotracks: A long-term egocentric visual object tracking dataset},
  author={Tang, Hao and Liang, Kevin J and Grauman, Kristen and Feiszli, Matt and Wang, Weiyao},
  journal=NIPS,
  volume={36},
  pages={75716--75739},
  year={2023}
}

@inproceedings{kwon2021h2o,
  title={H2o: Two hands manipulating objects for first person interaction recognition},
  author={Kwon, Taein and Tekin, Bugra and St{\"u}hmer, Jan and Bogo, Federica and Pollefeys, Marc},
  booktitle=ICCV,
  pages={10138--10148},
  year={2021}
}

@inproceedings{banerjee2025hot3d,
  title={HOT3D: Hand and Object Tracking in 3D from Egocentric Multi-View Videos},
  author={Banerjee, Prithviraj and Shkodrani, Sindi and Moulon, Pierre and Hampali, Shreyas and Han, Shangchen and Zhang, Fan and Zhang, Linguang and Fountain, Jade and Miller, Edward and Basol, Selen and others},
  booktitle=CVPR,
  pages={7061--7071},
  year={2025},
  organization={IEEE}
}

@article{damen2022rescaling,
  title={Rescaling egocentric vision: Collection, pipeline and challenges for epic-kitchens-100},
  author={Damen, Dima and Doughty, Hazel and Farinella, Giovanni Maria and Furnari, Antonino and Kazakos, Evangelos and Ma, Jian and Moltisanti, Davide and Munro, Jonathan and Perrett, Toby and Price, Will and others},
  journal=IJCV,
  volume={130},
  number={1},
  pages={33--55},
  year={2022},
  publisher={Springer}
}

@inproceedings{grauman2022ego4d,
  title={Ego4d: Around the world in 3,000 hours of egocentric video},
  author={Grauman, Kristen and Westbury, Andrew and Byrne, Eugene and Chavis, Zachary and Furnari, Antonino and Girdhar, Rohit and Hamburger, Jackson and Jiang, Hao and Liu, Miao and Liu, Xingyu and others},
  booktitle=CVPR,
  pages={18995--19012},
  year={2022}
}

@inproceedings{grauman2024ego,
  title={Ego-exo4d: Understanding skilled human activity from first-and third-person perspectives},
  author={Grauman, Kristen and Westbury, Andrew and Torresani, Lorenzo and Kitani, Kris and Malik, Jitendra and Afouras, Triantafyllos and Ashutosh, Kumar and Baiyya, Vijay and Bansal, Siddhant and Boote, Bikram and others},
  booktitle=CVPR,
  pages={19383--19400},
  year={2024}
}

@inproceedings{xu2025dexumi,
  title={DexUMI: Using Human Hand as the Universal Manipulation Interface for Dexterous Manipulation},
  author={Xu, Mengda and Zhang, Han and Hou, Yifan and Xu, Zhenjia and Fan, Linxi and Veloso, Manuela and Song, Shuran},
  booktitle=CORL,
  pages={437--459},
  year={2025},
  organization={PMLR}
}

@inproceedings{liu2025maniwav,
    title={ManiWAV: Learning Robot Manipulation from In-the-Wild Audio-Visual Data},
    author={Liu, Zeyi and Chi, Cheng and Cousineau, Eric and Kuppuswamy, Naveen and Burchfiel, Benjamin and Song, Shuran},
    booktitle=CORL,
    pages={947--962},
    year={2025},
    organization={PMLR}
}

@inproceedings{chi2024universal,
	title={Universal Manipulation Interface: In-The-Wild Robot Teaching Without In-The-Wild Robots},
	author={Chi, Cheng and Xu, Zhenjia and Pan, Chuer and Cousineau, Eric and Burchfiel, Benjamin and Feng, Siyuan and Tedrake, Russ and Song, Shuran},
	booktitle=RSS,
	year={2024}
}

@inproceedings{ha2024umionlegs,
title={{UMI}-on-Legs: Making Manipulation Policies Mobile with Manipulation-Centric Whole-body Controllers},
author={Huy Ha and Yihuai Gao and Zipeng Fu and Jie Tan and Shuran Song},
booktitle=CORL,
year={2024},
url={https://openreview.net/forum?id=3i7j8ZPnbm}
}

@inproceedings{lin2025data,
title={Data Scaling Laws in Imitation Learning for Robotic Manipulation},
author={Fanqi Lin and Yingdong Hu and Pingyue Sheng and Chuan Wen and Jiacheng You and Yang Gao},
booktitle=ICLR,
year={2025},
url={https://openreview.net/forum?id=pISLZG7ktL}
}

@article{nai2026humanoid,
  title={Humanoid Manipulation Interface: Humanoid Whole-Body Manipulation from Robot-Free Demonstrations},
  author={Nai, Ruiqian and Zheng, Boyuan and Zhao, Junming and Zhu, Haodong and Dai, Sicong and Chen, Zunhao and Hu, Yihang and Hu, Yingdong and Zhang, Tong and Wen, Chuan and others},
  journal={arXiv preprint arXiv:2602.06643},
  year={2026}
}

@misc{gupta2025umi,
      title={UMI-on-Air: Embodiment-Aware Guidance for Embodiment-Agnostic Visuomotor Policies}, 
      author={Harsh Gupta and Xiaofeng Guo and Huy Ha and Chuer Pan and Muqing Cao and Dongjae Lee and Sebastian Scherer and Shuran Song and Guanya Shi},
      year={2025},
      eprint={2510.02614},
      archivePrefix={arXiv},
      primaryClass={cs.RO},
      url={https://arxiv.org/abs/2510.02614}, 
}

@article{liu2025fastumi,
  title={FastUMI-100K: Advancing Data-driven Robotic Manipulation with a Large-scale UMI-style Dataset},
  author={Liu, Kehui and Jia, Zhongjie and Li, Yang and Chen, Pengan and Liu, Song and Liu, Xin and Zhang, Pingrui and Song, Haoming and Ye, Xinyi and Cao, Nieqing and others},
  journal={arXiv preprint arXiv:2510.08022},
  year={2025}
}

@article{seo2025legato,
  title={Legato: Cross-embodiment imitation using a grasping tool},
  author={Seo, Mingyo and Park, H Andy and Yuan, Shenli and Zhu, Yuke and Sentis, Luis},
  journal=RAL,
  volume={10},
  number={3},
  pages={2854--2861},
  year={2025},
  publisher={IEEE}
}

@article{tao2025dexwild,
  title={DexWild: Dexterous Human Interactions for In-the-Wild Robot Policies},
  author={Tao, Tony and Srirama, Mohan Kumar and Liu, Jason Jingzhou and Shaw, Kenneth and Pathak, Deepak},
  journal=RSS,
  year={2025}
}

@article{zhaxizhuoma2025fastumi,
  title={FastUMI: A Scalable and Hardware-Independent Universal Manipulation Interface with Dataset}, 
  author={Zhaxizhuoma and Kehui Liu and Chuyue Guan and Zhongjie Jia and Ziniu Wu and Xin Liu and Tianyu Wang and Shuai Liang and Pengan Chen and Pingrui Zhang and Haoming Song and Delin Qu and Dong Wang and Zhigang Wang and Nieqing Cao and Yan Ding and Bin Zhao and Xuelong Li},
  year={2025},
  eprint={2409.19499},
  journal={arXiv},
  primaryClass={cs.RO},
  url={https://arxiv.org/abs/2409.19499}, 
}

@inproceedings{black2025pi,
title={\${\textbackslash}pi\_\{0.5\}\$: a Vision-Language-Action Model with Open-World Generalization},
author={Kevin Black and Noah Brown and James Darpinian and Karan Dhabalia and Danny Driess and Adnan Esmail and Michael Robert Equi and Chelsea Finn and Niccolo Fusai and Manuel Y. Galliker and Dibya Ghosh and Lachy Groom and Karol Hausman and brian ichter and Szymon Jakubczak and Tim Jones and Liyiming Ke and Devin LeBlanc and Sergey Levine and Adrian Li-Bell and Mohith Mothukuri and Suraj Nair and Karl Pertsch and Allen Z. Ren and Lucy Xiaoyang Shi and Laura Smith and Jost Tobias Springenberg and Kyle Stachowicz and James Tanner and Quan Vuong and Homer Walke and Anna Walling and Haohuan Wang and Lili Yu and Ury Zhilinsky},
booktitle=CORL,
year={2025},
}

@article{team2025gemini,
  title={Gemini robotics 1.5: Pushing the frontier of generalist robots with advanced embodied reasoning, thinking, and motion transfer},
  author={Team, Gemini Robotics and Abdolmaleki, Abbas and Abeyruwan, Saminda and Ainslie, Joshua and Alayrac, Jean-Baptiste and Arenas, Montserrat Gonzalez and Balakrishna, Ashwin and Batchelor, Nathan and Bewley, Alex and Bingham, Jeff and others},
  journal={arXiv preprint arXiv:2510.03342},
  year={2025}
}

@inproceedings{zitkovich2023rt,
title={{RT}-2: Vision-Language-Action Models Transfer Web Knowledge to Robotic Control},
author={Brianna Zitkovich and Tianhe Yu and Sichun Xu and Peng Xu and Ted Xiao and Fei Xia and Jialin Wu and Paul Wohlhart and Stefan Welker and Ayzaan Wahid and Quan Vuong and Vincent Vanhoucke and Huong Tran and Radu Soricut and Anikait Singh and Jaspiar Singh and Pierre Sermanet and Pannag R Sanketi and Grecia Salazar and Michael S Ryoo and Krista Reymann and Kanishka Rao and Karl Pertsch and Igor Mordatch and Henryk Michalewski and Yao Lu and Sergey Levine and Lisa Lee and Tsang-Wei Edward Lee and Isabel Leal and Yuheng Kuang and Dmitry Kalashnikov and Ryan Julian and Nikhil J Joshi and Alex Irpan and brian ichter and Jasmine Hsu and Alexander Herzog and Karol Hausman and Keerthana Gopalakrishnan and Chuyuan Fu and Pete Florence and Chelsea Finn and Kumar Avinava Dubey and Danny Driess and Tianli Ding and Krzysztof Marcin Choromanski and Xi Chen and Yevgen Chebotar and Justice Carbajal and Noah Brown and Anthony Brohan and Montserrat Gonzalez Arenas and Kehang Han},
booktitle=CORL,
year={2023},
}

@inproceedings{octo_2024,
    title={Octo: An Open-Source Generalist Robot Policy},
    author = {{Octo Model Team} and Dibya Ghosh and Homer Walke and Karl Pertsch and Kevin Black and Oier Mees and Sudeep Dasari and Joey Hejna and Charles Xu and Jianlan Luo and Tobias Kreiman and {You Liang} Tan and Lawrence Yunliang Chen and Pannag Sanketi and Quan Vuong and Ted Xiao and Dorsa Sadigh and Chelsea Finn and Sergey Levine},
    booktitle = RSS,
    address  = {Delft, Netherlands},
    year = {2024},
}

@inproceedings{kim2024openvla,
title={Open{VLA}: An Open-Source Vision-Language-Action Model},
author={Moo Jin Kim and Karl Pertsch and Siddharth Karamcheti and Ted Xiao and Ashwin Balakrishna and Suraj Nair and Rafael Rafailov and Ethan P Foster and Pannag R Sanketi and Quan Vuong and Thomas Kollar and Benjamin Burchfiel and Russ Tedrake and Dorsa Sadigh and Sergey Levine and Percy Liang and Chelsea Finn},
booktitle=CORL,
year={2024},
}

@article{goyal2024rvt,
  author    = {Goyal, Ankit and Blukis, Valts and Xu, Jie and Guo, Yijie and Chao, Yu-Wei and Fox, Dieter},
  title     = {RVT2: Learning Precise Manipulation from Few Demonstrations},
  journal   = RSS,
  year      = {2024},
}

@inproceedings{huang2023voxposer,
title={VoxPoser: Composable 3D Value Maps for Robotic Manipulation with Language Models},
author={Wenlong Huang and Chen Wang and Ruohan Zhang and Yunzhu Li and Jiajun Wu and Li Fei-Fei},
booktitle=CORL,
year={2023},
}

@inproceedings{shridhar2023perceiver,
  title={Perceiver-actor: A multi-task transformer for robotic manipulation},
  author={Shridhar, Mohit and Manuelli, Lucas and Fox, Dieter},
  booktitle=CORL,
  pages={785--799},
  year={2023},
  organization={PMLR}
}

@inproceedings{shridhar2022cliport,
  title={Cliport: What and where pathways for robotic manipulation},
  author={Shridhar, Mohit and Manuelli, Lucas and Fox, Dieter},
  booktitle=CORL,
  pages={894--906},
  year={2022},
  organization={PMLR}
}

@inproceedings{lynch2021language,
  title   = {Language Conditioned Imitation Learning over Unstructured Data},
  author  = {Lynch, Corey and Sermanet, Pierre},
  booktitle = RSS,
  year    = {2021},
  url     = {https://arxiv.org/abs/2005.07648},
}

@inproceedings{yu2020meta,
  title={Meta-world: A benchmark and evaluation for multi-task and meta reinforcement learning},
  author={Yu, Tianhe and Quillen, Deirdre and He, Zhanpeng and Julian, Ryan and Hausman, Karol and Finn, Chelsea and Levine, Sergey},
  booktitle=CORL,
  pages={1094--1100},
  year={2020},
  organization={PMLR}
}

@inproceedings{yang2020multitask,
 author = {Yang, Ruihan and Xu, Huazhe and WU, YI and Wang, Xiaolong},
 booktitle = NIPS,
 editor = {H. Larochelle and M. Ranzato and R. Hadsell and M. F. Balcan and H. Lin},
 pages = {4767--4777},
 publisher = {Curran Associates, Inc.},
 title = {Multi-Task Reinforcement Learning with Soft Modularization},
 url = {https://proceedings.neurips.cc/paper/2020/file/32cfdce9631d8c7906e8e9d6e68b514b-Paper.pdf},
 volume = {33},
 year = {2020}
}

@inproceedings{luo2025grounding,
title={Grounding Video Models to Actions through Goal Conditioned Exploration},
author={Yunhao Luo and Yilun Du},
booktitle=ICLR,
year={2025},
url={https://openreview.net/forum?id=G6dMvRuhFr}
}

@inproceedings{reuss2023goal,
  title     = {Goal-conditioned imitation learning using score-based diffusion policies},
  author    = {Reuss, Moritz and Li, Maximilian and Jia, Xiaogang and Lioutikov, Rudolf},
  booktitle = RSS,
  year      = {2023},
  url       = {https://www.roboticsproceedings.org/rss19/p028.pdf}
}

@inproceedings{srinivas2018universal,
  title={Universal planning networks: Learning generalizable representations for visuomotor control},
  author={Srinivas, Aravind and Jabri, Allan and Abbeel, Pieter and Levine, Sergey and Finn, Chelsea},
  booktitle=ICML,
  pages={4732--4741},
  year={2018},
  organization={PMLR}
}

@article{nair2018visual,
  title={Visual reinforcement learning with imagined goals},
  author={Nair, Ashvin V and Pong, Vitchyr and Dalal, Murtaza and Bahl, Shikhar and Lin, Steven and Levine, Sergey},
  journal=NIPS,
  volume={31},
  year={2018}
}

@article{oquab2024dinov2,
  title={DINOv2: Learning Robust Visual Features without Supervision},
  author={Oquab, Maxime and Darcet, Timoth{\'e}e and Moutakanni, Th{\'e}o and Vo, Huy V. and Szafraniec, Marc and Khalidov, Vasil and Fernandez, Pierre and Haziza, Daniel and Massa, Francisco and El-Nouby, Alaaeldin and Assran, Mahmoud and Ballas, Nicolas and Galuba, Wojciech and Howes, Russell and Huang, Po-Yao and Li, Shang-Wen and Misra, Ishan and Rabbat, Michael and Sharma, Vasu and Synnaeve, Gabriel and Xu, Hu and Jegou, Herv{\'e} and Mairal, Julien and Labatut, Patrick and Joulin, Armand and Bojanowski, Piotr},
  journal=TMLR,
  year={2024}
}

@article{mildenhall2021nerf,
  title={Nerf: Representing scenes as neural radiance fields for view synthesis},
  author={Mildenhall, Ben and Srinivasan, Pratul P and Tancik, Matthew and Barron, Jonathan T and Ramamoorthi, Ravi and Ng, Ren},
  journal={Communications of the ACM},
  volume={65},
  number={1},
  pages={99--106},
  year={2021}
}

@inproceedings{tancik2020fourier,
  title={Fourier features let networks learn high frequency functions in low dimensional domains},
  author={Tancik, Matthew and Srinivasan, Pratul and Mildenhall, Ben and Fridovich-Keil, Sara and Raghavan, Nithin and Singhal, Utkarsh and Ramamoorthi, Ravi and Barron, Jonathan and Ng, Ren},
  booktitle=NIPS,
  volume={33},
  pages={7537--7547},
  year={2020}
}

@inproceedings{vaswani2017attention,
  title={Attention is all you need},
  author={Vaswani, Ashish and Shazeer, Noam and Parmar, Niki and Uszkoreit, Jakob and Jones, Llion and Gomez, Aidan N and Kaiser, {\L}ukasz and Polosukhin, Illia},
  booktitle=NIPS,
  volume={30},
  year={2017}
}

@inproceedings{ho2020denoising,
  title={Denoising diffusion probabilistic models},
  author={Ho, Jonathan and Jain, Ajay and Abbeel, Pieter},
  booktitle=NIPS,
  volume={33},
  pages={6840--6851},
  year={2020}
}

@inproceedings{lipmanflow,
  title={Flow Matching for Generative Modeling},
  author={Lipman, Yaron and Chen, Ricky TQ and Ben-Hamu, Heli and Nickel, Maximilian and Le, Matthew},
  booktitle=ICLR,
  year={2020}
}
}

\clearpage
\appendix
\maketitlesupplementary
\section{Dataset Documentation}
\begin{figure}
    \centering
    \includegraphics[width=0.95\linewidth]{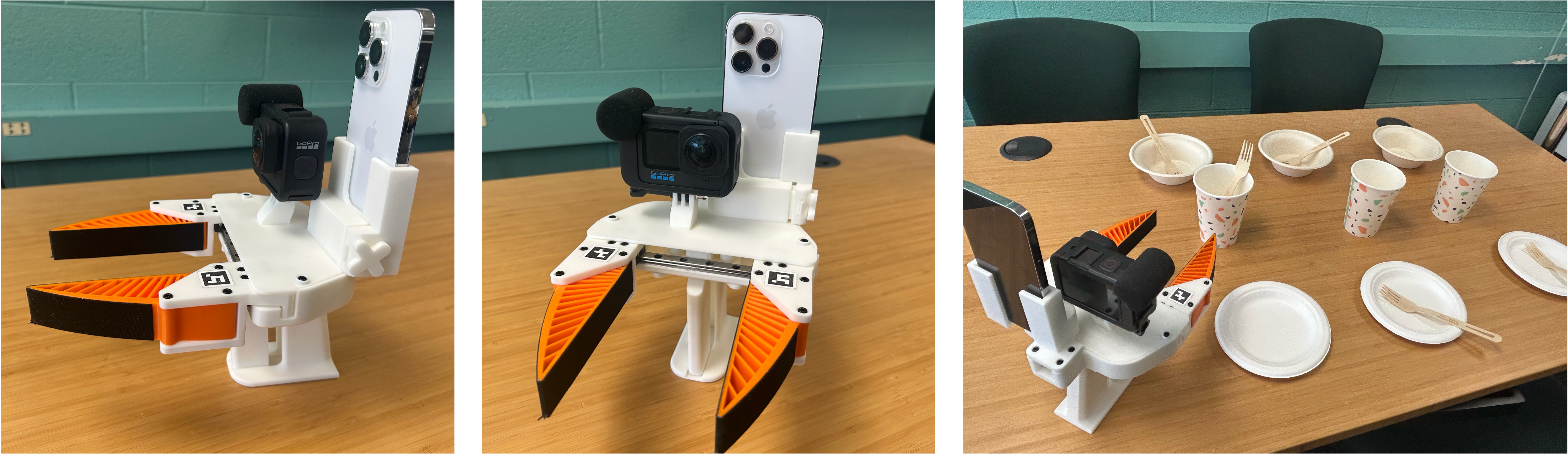}
    \caption{More visualization demonstrations of modified UMI device.}
    \label{fig:modified-umi}
\end{figure}

\textbf{Collection setup.}
EgoSPT contains 2,841 egocentric pick-and-place episodes collected with a modified UMI (see Fig. \ref{fig:modified-umi}) by nine trained experts. 
Each episode records an RGB video and tracking streams. The recording interface stores raw episode videos as \texttt{camera\_<i>.mp4} 
and per-camera arrays such as pose, receive time, capture time, and intrinsics. In the processed dataset used by SPOT, \texttt{camera\_1.mp4} is the RGB video consumed by the vision model, 
while the tracking stream is used to recover time-aligned 6-DoF EE motion. 
Videos are recorded at 1920$\times$1080 and 30 fps, and are temporally subsampled with stride 3 for trajectory prediction.

\textbf{Trajectory processing pipeline.}
The raw demonstrations are converted into the format consumed by SPOT data loader. 
The practical pipeline is: record raw videos and tracking arrays, synchronize RGB and tracking timestamps, interpolate EE pose onto valid RGB frame times, detect ArUco tags on the gripper, calibrate and interpolate gripper width, 
copy the RGB video into a processed episode directory, and attach first-frame object/target bounding-box annotations. The processed layout expected by the dataset loader is
\begin{center}
\path{<data.root>/<scene>/<task>/recording_output_processed/<episode>/}
\end{center}
containing \texttt{camera\_1.mp4}, \texttt{valid\_indices}, \texttt{pose\_interp}, and \texttt{gripper\_widths}. 
The raw videos and processed trajectories are organized under \texttt{umi\_day/output\_data}, and first-frame prompt annotations are stored separately in \texttt{annotations\_merged.json}.

\textbf{Time alignment and trajectory recovery.}
During synchronization, camera and tracking receive times are corrected using configured camera and tracking latencies. The overlapping time range is used to define \texttt{valid\_indices} over RGB frames. EE translations are linearly interpolated, rotations are interpolated with spherical linear interpolation, and the camera pose is converted to an EE pose using the calibrated camera-to-EE transform. Gripper width is obtained from ArUco detections of finger tags, calibrated with a gripper-range episode, and interpolated to the valid frame times. The resulting \texttt{pose\_interp} and \texttt{gripper\_widths} are aligned with the valid RGB frames.

\begin{figure}[t]
    \centering
    \includegraphics[width=0.95\linewidth]{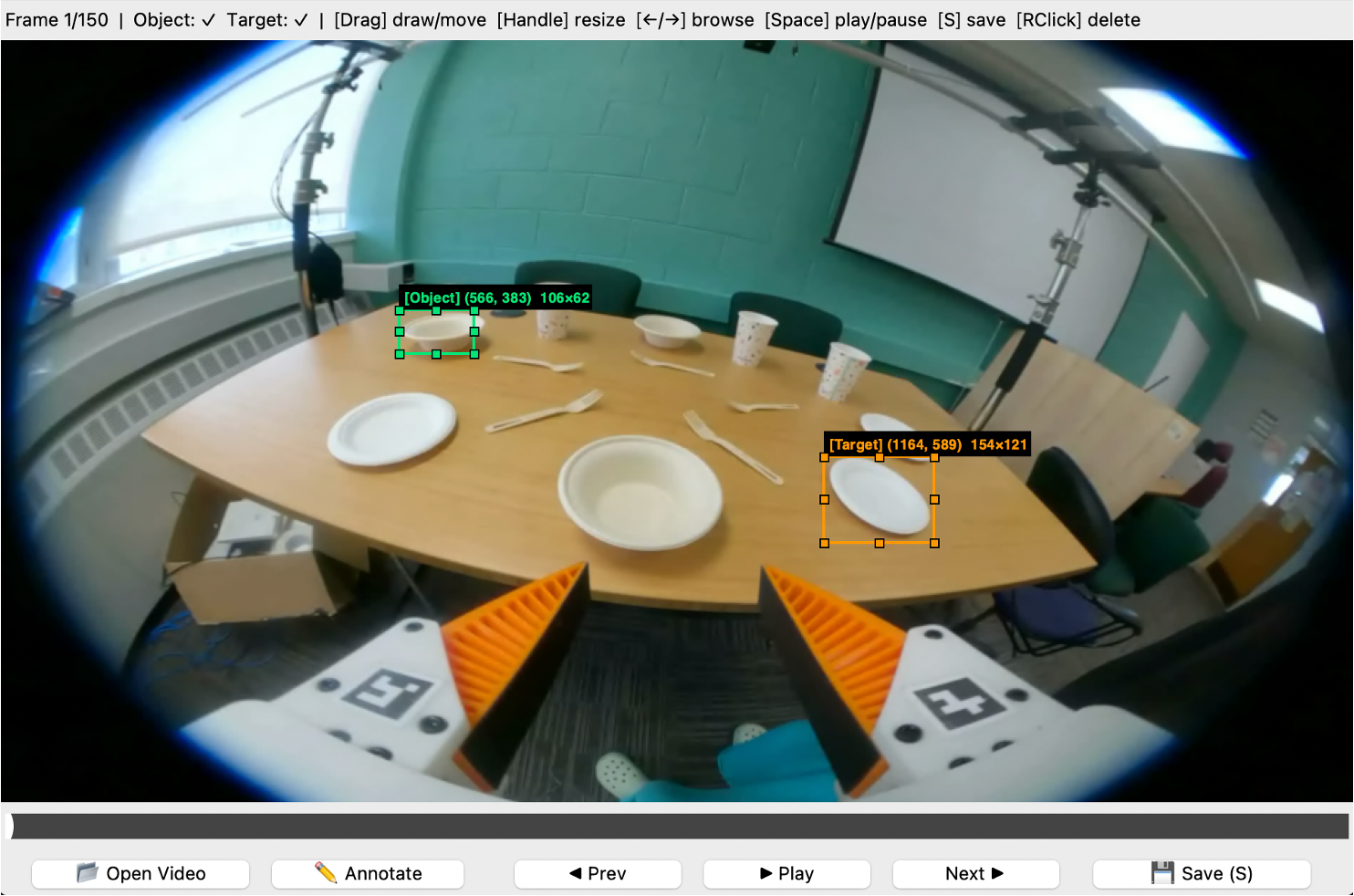}
    \caption{\textbf{Annotation interface.} Annotators label the manipulated object and target region on the first egocentric frame. The resulting bounding boxes are stored in the annotation JSON and used as spatial prompts for SP-VTP.}
    \label{fig:annotation-tool}
\end{figure}

\textbf{Annotations and sample construction.}
Nine trained experts are assigned the collected videos and annotate the object and target bounding boxes using our annotation tool, as shown in Fig.~\ref{fig:annotation-tool}.
The tool is a lightweight video bounding-box annotator designed for frame-0 task specification: annotators open an episode video, jump to the first frame, and draw exactly two boxes, where the first box denotes the manipulated object and the second denotes the target region.
It also supports video browsing, playback, zooming, box editing, and in-place correction of merged annotations.
Annotations are saved in JSON format with both the ordered box list and explicit \texttt{object}/\texttt{target} fields in original video pixel coordinates. The annotation results are merged and stored in \texttt{annotations\_merged.json}.

\begin{figure}[t]
    \centering
    \includegraphics[width=0.95\linewidth]{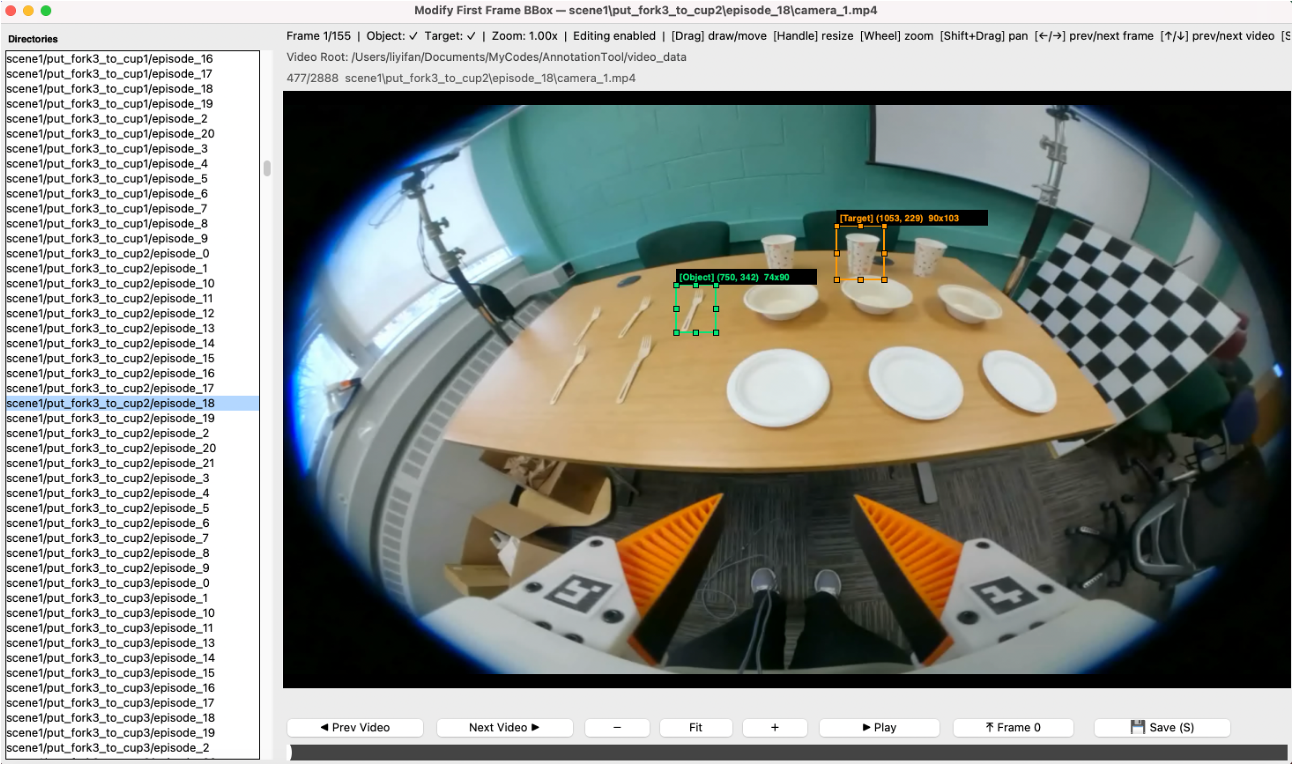}
    \caption{\textbf{Annotation modification interface.} After initial annotation, annotators inspect each video and correct frame-0 object/target bounding boxes directly in the merged annotation file.}
    \label{fig:annotation-modify-tool}
\end{figure}

After annotation, we manually check each bounding box with a separate annotation modification tool shown in Fig.~\ref{fig:annotation-modify-tool}. This tool loads videos from \texttt{annotations\_merged.json}, supports direct navigation through annotated videos, allows frame browsing and playback, and restricts box editing to frame 0. Annotators can move, resize, create, or delete boxes, and the corrected object/target boxes are written back to the original merged annotation file in place.

The annotation JSON maps each episode video key to the object and target boxes in the original first-frame pixel coordinates. The dataset parser converts keys into \texttt{scene}, \texttt{task}, and \texttt{episode}, locates the corresponding processed episode directory, and skips episodes with missing videos, missing prompts, missing zarr arrays, or too few valid frames. For a sample at timestep $t$, the loader reads the first frame, the current RGB frame, normalized object/target prompts, a history window, and a future action window. Future actions are computed as relative EE motions in the current frame:
\[
T_{\mathrm{rel}}(t,h)=T_t^{-1}T_{t+h},
\]
then represented as relative translation, 6D rotation, and gripper width. Action history is built analogously from previous poses relative to the current pose. Each returned sample contains \texttt{first\_frame}, \texttt{current\_frame}, \texttt{prompt\_object}, \texttt{prompt\_target}, \texttt{action\_history}, \texttt{future\_actions}, and \texttt{is\_final\_chunk}.

\textbf{Preprocessing and scale.}
The default preprocessing uses image size $224\times224$, bounding-box prompts, action horizon $H=16$, history horizon $K=4$, and stride 3. This produces 112,856 sliding-window trajectory samples in total, including 102,832 training samples from 2,584 episodes and 10,024 validation samples from 257 episodes. The mean episode length after preprocessing is 59.7 trajectory steps, and each episode yields 39.7 trajectory samples on average. Across all samples, the mean relative-translation magnitude is $21.59$ cm, the mean final displacement is $35.29$ cm, and the mean gripper width is $30.3$ mm.

\textbf{Scene and task coverage.}
The data cover five visually similar forks and nine target receptacles across three scenes. Scene 1 uses structured layouts, Scene 2 uses a cluttered layout, and Scene 3 uses diverse cluttered subscenes with fewer demonstrations per configuration. The full dataset contains 931 episodes / 43,972 samples from Scene 1, 919 episodes / 35,902 samples from Scene 2, and 991 episodes / 32,982 samples from Scene 3. Scene 1 and Scene 2 are organized around object--target tasks such as \texttt{put\_fork1\_to\_plate1}; Scene 3 is organized into 22 cluttered subscenes, each covering the object--target combinations with sparse demonstrations. The train/validation split is performed at the episode/task level while preserving scene coverage; the scene-distribution total variation between splits is 0.0015. All reported experiments therefore evaluate both overall validation performance and per-scene behavior.

\textbf{Quality checks and statistics files.}
Before training, we verify that every annotation key maps to an existing processed episode, each processed episode has \texttt{camera\_1.mp4}, \texttt{valid\_indices}, \texttt{pose\_interp}, and \texttt{gripper\_widths}, each episode has enough valid frames for $K+H+1$, the pose and gripper arrays share compatible temporal indexing, and first-frame boxes are valid in the original image coordinate system. We also check that scene/task/episode names in the annotations match directory names after path normalization and that the gripper calibration and tag detections are available for width interpolation.

We generate auxiliary statistics files to support reproducibility and data inspection. \texttt{summary.json} stores the machine-readable dataset summary, \texttt{episode\_stats.csv} stores per-episode lengths, sample counts, and video metadata, \texttt{count\_stats.csv} stores scene/task/scene-task counts, \texttt{prompt\_stats.csv} stores bounding-box size and location statistics, \texttt{image\_stats.csv} stores sampled RGB statistics, and \texttt{outliers\_*.csv} stores samples with the largest final-displacement values for train and validation splits. These files are used to verify dataset scale, split quality, prompt validity, and trajectory target ranges. These files are provided in the supplementary material.

\section{Model Architecture Details}
\textbf{Policy interface.}
The implementation is centered on \texttt{VisionTrajPolicy}, which maps a first-frame task specification and a current execution state to a future action chunk:
\[
(\text{first frame}, \text{prompt}, \text{current frame}, \text{history})
\rightarrow A_t \in \mathbb{R}^{H\times10}.
\]
We use $H=16$ by default. Each action vector is $[\Delta x,\Delta y,\Delta z,\mathrm{rot6d},g]$, where the pose component is expressed relative to the current camera/EE frame and $g$ is the gripper width. Architecturally, the policy has three parts: the task encoder builds prompt-conditioned first-frame tokens, the observation encoder builds current-state tokens, and the action head decodes future trajectory tokens from their concatenation.

\textbf{Prompt forms.}
The dataset loader exposes seven task-input variants. \texttt{bbox} uses a raw first frame with object/target bounding-box coordinates; \texttt{point} replaces each box with its center; \texttt{none} removes coordinate prompts and uses learned null prompt tokens; \texttt{vision\_bbox} renders object/target boxes on the first frame without coordinate prompts; and \texttt{vision\_bbox\_and\_bbox} combines rendered boxes with coordinate tokens. \texttt{object\_only} and \texttt{target\_only} retain only the corresponding rendered box and coordinate prompt. The final SPOT configuration uses \texttt{vision\_bbox\_and\_bbox}, so the task is represented in both image space and coordinate-token form.

\textbf{Shared visual backbone.}
Both the task encoder and observation encoder use the same visual backbone interface. The backbone is implemented with \texttt{timm} and returns image tokens in $(B,N,C)$ format. The default backbone is \texttt{vit\_base\_patch14\_dinov2}, with $C=768$ and policy dimension $D=768$. We also support SAM-Base, Perception Encoder-Base, SigLIP-Base, and EVA2-Base through their corresponding \texttt{timm} model names. When the backbone is frozen, its parameters are excluded from the optimizer and the module is kept in evaluation mode; when unfrozen, it is trained together with the policy.

\textbf{Task encoder.}
The task encoder first applies the shared visual backbone to the first frame, optionally after visual prompt rendering. Image tokens are projected to dimension $D$ and receive image type embeddings. Coordinate prompts are encoded separately: a bounding box contributes two corner tokens per retained role, a point prompt contributes one token per retained role, and the no-prompt setting uses two learned null tokens. Prompt tokens receive role/type embeddings before fusion.

The default fusion path is prompt-to-image cross-attention. Prompt tokens query first-frame image tokens through a Transformer decoder with 4 layers and 12 attention heads. With image-summary return enabled, the task encoder returns the first image summary token followed by the fused prompt tokens. Thus, the default bounding-box coordinate prompt produces five task tokens: one image-summary token and four fused coordinate-prompt tokens.

\textbf{Observation encoder and conditioning sequence.}
The observation encoder applies the same image encoder and projection layer to the current frame, giving current-frame image tokens in the same feature space as the first-frame task tokens. When the history horizon $K>0$, each previous 10D action is embedded by a lightweight MLP, \texttt{Linear(10,D)-GELU-Linear(D,D)}, then augmented with learnable temporal position embeddings and a history type embedding. The observation sequence is the concatenation of current-frame image tokens and history-action tokens.

The final conditioning sequence is
\[
Z_{\mathrm{cond}}=[Z_{\mathrm{task}};Z_{\mathrm{obs}}],
\]
which contains first-frame task evidence, current egocentric visual context, and recent motion context. The trajectory head uses this sequence as cross-attention memory while decoding the future action chunk.

\section{Training and Evaluation Details}
\subsection{Training Details}
\textbf{Default configuration.}
The SPOT prompting baselines and architecture ablations change one factor at a time from the default configuration. The default model uses \texttt{vision\_bbox\_and\_bbox} prompts, a frozen DINOv2 ViT-B/14 visual encoder, policy dimension $D=768$, cross-attention task fusion with 4 layers and 12 heads, history horizon $K=4$, a 6-layer 12-head trajectory decoder, and a flow-matching head. The three-seed prompting runs use learning rate $2\times10^{-4}$, effective global batch size 128, 30 epochs, validation every 2 epochs, 4 dataloader workers, and bf16 mixed precision. To improve data-loading efficiency while preserving the global batch size, seed 42 uses 8 GPUs with 16 samples per GPU, seed 0 uses 2 GPUs with 64 samples per GPU, and seed 1 uses one GPU with batch size 128. Predecoded frame caching changes only I/O and was verified to return tensors identical to direct video decoding.

\textbf{Split construction.}
The training split is deterministic. Episodes are grouped by scene and task, task names are shuffled with a seed-dependent scene-specific random generator, and validation tasks are selected according to the validation ratio. All episodes of a selected validation task are assigned to validation, preventing sample-level leakage across the same task split decision. Rank 0 writes a \texttt{split\_manifest.json} containing train/validation episode keys, scene/task names, and paths, and training checks that no episode key appears in both splits.

\textbf{Optimization.}
Training uses \texttt{accelerate} for distributed execution. The optimizer is AdamW over parameters with \texttt{requires\_grad=True}; thus frozen visual backbones are excluded from optimization. Gradients are clipped to norm 1.0 when synchronized. The learning-rate scheduler is constructed with \texttt{diffusers.optimization.get\_scheduler}, and the code supports constant, warmup, linear, cosine, cosine-with-restarts, and polynomial schedules.

\textbf{Action normalization.}
Before optimizer construction, the policy computes per-action-dimension mean and standard deviation from training samples. Standard deviations are clamped to at least $10^{-4}$. During training, future action chunks are normalized as $(a-\mu)/\sigma$ before being passed to the trajectory head; predictions are denormalized before evaluation or visualization.

\textbf{Flow-matching head.}
The default flow-matching head predicts velocity over normalized action chunks. During training, time $t$ is sampled from a Beta distribution with $\alpha=1.5$ and $\beta=1.0$, then clamped away from zero by $t=0.999t+0.001$. Given a clean action chunk $x_0$ and Gaussian noise $\epsilon$, the interpolated sample is $x_t=(1-t)x_0+t\epsilon$, and the target velocity is $\epsilon-x_0$. Time is encoded with a continuous sinusoidal embedding over periods from 0.004 to 4.0, followed by a two-layer MLP.

\textbf{Diffusion head.}
The alternative diffusion head uses a DDPM scheduler with the \texttt{squaredcos\_cap\_v2} beta schedule during training. It predicts either noise or the clean sample depending on the configured prediction type. The diffusion and flow-matching heads share the same high-level decoder structure: trajectory tokens with a time embedding attend to the conditioning tokens through a Transformer decoder and output an action chunk.

\subsection{Evaluation Details}
\textbf{Validation protocol.}
Training-time validation runs every two epochs on a capped number of batches for fast feedback. It reports both overall metrics and per-scene metrics using validation episodes selected at training start. Final evaluation reuses the same validation episodes and evaluates the full validation set without batch truncation. Each offline sample receives its recorded current RGB frame and recorded $K=4$ action history; a predicted chunk is not fed back as the history of a later sample. Ground-truth future actions are used only as evaluation targets.

\textbf{Inference.}
For the default flow-matching head, inference starts from Gaussian noise at $t=1$ and integrates to $t=0$ using forward Euler with 10 steps. The final evaluation uses deterministic initial action noise with sample seed 0. For the diffusion head, inference uses a DDIM scheduler with the same \texttt{squaredcos\_cap\_v2} beta schedule used during training.

\textbf{Metrics and outputs.}
Let $N$ denote the number of evaluated action chunks and $H=16$ their horizon. For sample $i$ and future step $h$, the prediction and target contain translation $\hat{p}_{i,h},p_{i,h}\in\mathbb{R}^3$, 6D rotation $\hat{r}_{i,h},r_{i,h}\in\mathbb{R}^6$, and gripper width $\hat{g}_{i,h},g_{i,h}$. We first define the per-waypoint translation and gripper errors as
\[
e^p_{i,h}=\lVert\hat{p}_{i,h}-p_{i,h}\rVert_2,
\qquad
e^g_{i,h}=|\hat{g}_{i,h}-g_{i,h}|.
\]
The reported mean position error is $\mathrm{Pos.~L2}=\frac{1}{NH}\sum_{i,h}e^p_{i,h}$, whereas final displacement error uses only the last waypoint, $\mathrm{FDE}=\frac{1}{N}\sum_i e^p_{i,H}$. Grip.~L1 is $\frac{1}{NH}\sum_{i,h}e^g_{i,h}$. Native trajectory values are stored in meters; we multiply translation errors by 100 for centimeters and gripper-width errors by 1000 for millimeters.

For orientation, each 6D vector is split into two 3D vectors $(a,b)$ and mapped to $\mathrm{SO}(3)$ by Gram--Schmidt: $u=a/\lVert a\rVert_2$, $v=(b-(u^\top b)u)/\lVert b-(u^\top b)u\rVert_2$, and $R=[u,v,u\times v]$. Given predicted and target matrices $\hat{R}_{i,h}$ and $R_{i,h}$, their geodesic error is
\[
e^R_{i,h}=\frac{180}{\pi}\cos^{-1}\!\left(\operatorname{clip}\!\left(\frac{\operatorname{tr}(\hat{R}_{i,h}^{\top}R_{i,h})-1}{2},-1,1\right)\right).
\]
Rot.~geo. averages $e^R_{i,h}$ over all $N H$ waypoints; endpoint Rot.~geo. averages $e^R_{i,H}$. R@$\tau$ is the percentage of chunks satisfying $e^R_{i,H}\leq\tau^\circ$ for $\tau\in\{10,20\}$. Rot6D L2, shown only in diagnostic tables, is the dimensionless quantity $\frac{1}{NH}\sum_{i,h}\lVert\hat{r}_{i,h}-r_{i,h}\rVert_2$ and is not used in the combined criterion.

Endpoint translation success S@$X$ is $\frac{100}{N}\sum_i\mathbb{1}[e^p_{i,H}\leq X]$ for $X\in\{2,5,10\}$ cm. We binarize gripper width as open when $g>30$ mm and closed otherwise. G-end is the percentage of chunks for which the predicted and target endpoint states agree. Combined@$X$ is the percentage satisfying all three conditions simultaneously:
\[
\begin{aligned}
e^p_{i,H}&\leq X, & e^R_{i,H}&\leq20^\circ,\\
\mathbb{1}[\hat{g}_{i,H}>30\text{ mm}]&=
\mathbb{1}[g_{i,H}>30\text{ mm}].
\end{aligned}
\]
with $X\in\{5,10\}$ cm. Unless a phase table explicitly states episode-balanced aggregation, overall metrics are means over all 10,024 validation chunks. Three-seed tables first compute each seed on the same samples and then report the arithmetic mean and sample standard deviation across seeds. All success measures are offline prediction criteria rather than robot task-success rates. Final evaluation writes overall and per-scene summaries to \texttt{eval\_metrics/all/summary.json}, \texttt{eval\_metrics/all/metrics.csv}, and the corresponding scene-specific files. Full-episode visualizations additionally stitch predicted chunks in temporal order to inspect accumulated drift.

\subsection{Additional Information-Source Controls}
\begin{table}[t]
  \centering
  \caption{Seed-42 information-source control on the same 10,024 validation samples. No Prompt retains the unmarked first frame and $K=4$ recorded action history, whereas Current RGB Only removes both task input and history. Grip.~L1 is in millimeters.}
  \label{tab:information-source-controls}
  \scriptsize
  \setlength{\tabcolsep}{2.5pt}
  \resizebox{\columnwidth}{!}{%
    \begin{tabular}{lccccc}
      \toprule
      Input & Pos.~L2 & FDE & Rot6D L2 & Grip.~L1 & S@2/5/10 (\%) \\
      \midrule
      No prompt & 9.12 & 17.39 & 0.2021 & 10.98 & 5.01/20.66/40.22 \\
      Current RGB only & 10.84 & 19.79 & 0.2133 & 13.93 & 3.72/17.94/36.24 \\
      \bottomrule
    \end{tabular}%
  }
\end{table}

\begin{table}[t]
  \centering
  \caption{Seed-42 observation-encoder control. Both rows retain the SPOT task encoder, prompt, $K=4$ history, action head, split, and evaluation samples. The second row replaces only the current-frame observation encoder with the SigLIP tower used by $\pi$0.5. Translation errors and Grip.~L1 are in centimeters.}
  \label{tab:observation-encoder-control}
  \scriptsize
  \setlength{\tabcolsep}{2.5pt}
  \resizebox{\columnwidth}{!}{%
    \begin{tabular}{lccccc}
      \toprule
      Observation encoder & Pos.~L2 & FDE & Rot6D L2 & Grip.~L1 & S@2/5/10 (\%) \\
      \midrule
      DINOv2 (default) & 6.99 & 11.47 & 0.1965 & 1.13 & 3.35/20.83/52.80 \\
      $\pi$0.5 SigLIP & 7.10 & 11.54 & 0.1969 & 1.33 & 3.93/20.81/50.96 \\
      \bottomrule
    \end{tabular}%
  }
\end{table}

Table~\ref{tab:information-source-controls} shows that the current RGB frame alone does not explain the No Prompt result: removing both the unmarked first frame and recorded history raises FDE by 2.40 cm and lowers S@10 by 3.98 percentage points. Because this intervention removes two sources together, it establishes their joint contribution but does not attribute the difference to either source individually. Table~\ref{tab:observation-encoder-control} further shows that replacing only the current-frame observation encoder with the $\pi$0.5 SigLIP tower leaves translation and rotation nearly unchanged and increases gripper error. Thus, the corresponding external-baseline behavior cannot be attributed to the observation tower alone.

\subsection{External VLA Baseline Protocols}
\label{sec:vla-protocols}
All external VLA rows use the same train/validation split, 16-step 10D action target, 10,024 validation samples, and training seeds 42/0/1. SmolVLA, OpenVLA-OFT, GR00T N1.6, and $\pi$0.5 use the current RGB frame, the same recorded 4-step action history as SPOT, and category-level language. They do not receive the first frame, rendered boxes, or box coordinates. Both object and target indices are removed from the instructions reported in Table~\ref{tab:external-baselines}, yielding prompts such as \texttt{pick fork and place to plate}.

We fine-tune the $\pi$0.5 action expert only. For SmolVLA we train the action expert and state projection while freezing the vision/VLM backbone; OpenVLA-OFT uses rank-32 LoRA and its continuous OFT L1 action head. Consequently, Table~\ref{tab:external-baselines} is an external reference under matched data, language-only task specification, and action targets, rather than an isolated architecture comparison.

\subsection{External VLA Input Diagnostics}
\begin{table}[t]
  \centering
  \caption{Inference-time VLA language sensitivity. Indexed instructions retain both instance IDs; Object-ID removed retains the target ID; Category removes both IDs and corresponds to the protocol reported in Table~\ref{tab:external-baselines}. Values are three-seed means; checkpoints and non-language inputs are fixed within each model.}
  \label{tab:vla-instance-id-diagnostic}
  \scriptsize
  \setlength{\tabcolsep}{2.2pt}
  \resizebox{\columnwidth}{!}{%
    \begin{tabular}{llcc}
      \toprule
      Model & Instruction & FDE (cm) & S@10 (\%) \\
      \midrule
      \multirow{3}{*}{SmolVLA}
        & Indexed & 12.66 & 49.05 \\
        & Object-ID removed & 13.98 & 42.68 \\
        & Category & 16.85 & 37.21 \\
      \midrule
      \multirow{3}{*}{OpenVLA-OFT}
        & Indexed & 11.11 & 56.68 \\
        & Object-ID removed & 12.38 & 48.95 \\
        & Category & 15.33 & 41.43 \\
      \midrule
      \multirow{3}{*}{GR00T N1.6}
        & Indexed & 12.77 & 48.08 \\
        & Object-ID removed & 15.42 & 35.38 \\
        & Category & 17.39 & 32.42 \\
      \bottomrule
    \end{tabular}%
  }
\end{table}

\begin{table}[t]
  \centering
  \caption{Seed-42 OpenVLA-OFT history control under indexed task language. The no-history model removes the 40D state/history projector and is trained and evaluated on the same sample windows. Grip.~L1 is in millimeters. This diagnostic is separate from the category-language comparison in Table~\ref{tab:external-baselines}.}
  \label{tab:openvla-history-diagnostic}
  \scriptsize
  \setlength{\tabcolsep}{2.4pt}
  \resizebox{\columnwidth}{!}{%
    \begin{tabular}{lccccc}
      \toprule
      History & Pos.~L2 & FDE & Rot6D L2 & Grip.~L1 & S@2/5/10 (\%) \\
      \midrule
      $K=4$ & 6.16 & 11.01 & 0.1657 & 7.98 & 5.38/28.21/57.74 \\
      $K=0$ & 6.68 & 11.33 & 0.1749 & 13.83 & 4.97/27.16/54.79 \\
      \bottomrule
    \end{tabular}%
  }
\end{table}

Table~\ref{tab:vla-instance-id-diagnostic} isolates the task information carried by language instance indices. Removing only the object index degrades all three language-conditioned VLAs, and removing both indices degrades them further. The category instruction is under-specified when several visually similar forks and receptacles are present, so this result measures lost task information rather than an architectural defect. Table~\ref{tab:openvla-history-diagnostic} separately shows that history has a modest effect on OpenVLA-OFT translation but a large effect on gripper prediction: removing history increases Grip.~L1 from 7.98 to 13.83 mm. This diagnostic uses indexed language and is therefore not treated as another row in the category-language comparison.

\subsection{Five-Stage Phase Segmentation}
\label{sec:phase-segmentation}
The analysis in Table~\ref{tab:phase-analysis} uses ground-truth gripper width and EE translation only to assign post-hoc evaluation groups; these signals are not additional policy inputs. For each episode, valid frames are subsampled with stride 3. We median-filter the gripper width and estimate its open level as the median over the first 12\% of the episode (at least five steps) and its closed level as the episode-wide 10th percentile. With normalized openness $q_t=(w_t-w_{\mathrm{closed}})/(w_{\mathrm{open}}-w_{\mathrm{closed}})$, closure completion is the first sustained run of two to five steps with $q_t\leq0.25$ between 8\% and 78\% episode progress. Grasp begins immediately after the last $q_t\geq0.75$ sample within the preceding 20\% of the episode. Pre-grasp begins approximately 10\% of an episode before grasp onset and is refined within a five-step neighborhood to the minimum EE translation speed. Transport begins after closure completion plus a short initial-lift interval of 5\% of the episode (at least two steps).

Release completion is the first two-step run with $q_t\geq0.40$ after 55\% episode progress; release onset follows the last $q_t\leq0.20$ sample in the preceding 20\% window. If no sustained reopening is found, we use the strongest late positive gripper-width derivative. Placement begins approximately 18\% of an episode before release onset and is refined within a seven-step neighborhood to the strongest EE deceleration, so it includes final target approach rather than release alone. The resulting intervals are Approach (episode start to pre-grasp), Pre-grasp (pre-grasp to closure onset), Grasp/Occlusion (closure onset through closure and initial lift), Transport (post-lift transfer), and Placement (final target approach through release and episode end). ``Occlusion'' is a semantic name for the grasp interval; visibility is not detected from images.

If an episode is too short, has less than 15 mm of calibrated gripper motion, or lacks reliable closure/release transitions, we use fixed progress boundaries at 28\%, 38\%, 52\%, and 78\%. The signal-based detector succeeds for 217 of 257 validation episodes and this fallback handles the remaining 40. We enforce ordered boundaries with at least two steps in each transition-centered phase; manual inspection of 15 episodes sampled across all scenes found no phase-order inversions. Per-timestep errors are assigned using the phase of each future target timestep, whereas FDE and other endpoint metrics are assigned using the phase of the 16-step action-chunk endpoint rather than that of the current input frame.

\subsection{Pick/Place Boundary Analysis}
\label{sec:pick-place-segmentation}
Table~\ref{tab:pick-place-analysis} uses an independent contact detector based on ground-truth EE translation and gripper width, again only for post-hoc grouping. We smooth the 3D EE trajectory and gripper signal, calibrate episode-specific open and closed levels, and identify sustained closing and reopening intervals. Within temporal neighborhoods around these transitions, the pick and place contacts are refined to the low-speed center of the lowest EE-height basin, using world $Y$ as the upward axis. The highest EE position between the two contacts defines the transport apex. If the gripper signal cannot be calibrated or produces inconsistent contact ordering, position-only low-height basins provide an explicit fallback.

The \emph{task-half} definition assigns a chunk to Pick when its future endpoint precedes the transport apex and to Place otherwise. The \emph{contact-window} definition retains endpoints within $\pm15$ raw video frames (approximately $\pm0.5$ s) of either detected contact and assigns an endpoint to the nearer contact. FDE and S@2/5/10 therefore correspond exactly to the classified endpoint, and results are first averaged within episode and then across episodes. Detector confidence combines contact-basin prominence, low EE speed, gripper/position agreement, gripper-motion amplitude, and plausible temporal ordering. The full analysis uses all 257 validation episodes; Table~\ref{tab:pick-place-high-confidence} reports the stricter 183-episode high-confidence subset.

\begin{table}[t]
  \centering
  \caption{Pick/place endpoint results on the 183-episode high-confidence boundary subset. FDE is in centimeters and S@10 is in percent. The conclusions match the all-episode results in Table~\ref{tab:pick-place-analysis}.}
  \label{tab:pick-place-high-confidence}
  \scriptsize
  \setlength{\tabcolsep}{2.2pt}
  \resizebox{\columnwidth}{!}{%
    \begin{tabular}{llcccc}
      \toprule
      Split & Condition & Pick FDE & Place FDE & Pick S@10 & Place S@10 \\
      \midrule
      \multirow{3}{*}{Task half}
        & No prompt     & 10.63 & 26.59 & 55.9 & 16.6 \\
        & Target only   & 10.72 & \textbf{11.75} & 53.5 & \textbf{47.2} \\
        & BBox + visual & \textbf{9.73} & 13.68 & \textbf{60.4} & 38.7 \\
      \midrule
      \multirow{3}{*}{Contact $\pm15$}
        & No prompt     & 8.08 & 24.99 & 70.8 & 25.6 \\
        & Target only   & 9.42 & \textbf{10.69} & 61.9 & \textbf{53.9} \\
        & BBox + visual & \textbf{7.62} & 13.19 & \textbf{75.2} & 43.8 \\
      \bottomrule
    \end{tabular}%
  }
\end{table}

Under both boundary definitions, BBox+Visual remains strongest for Pick FDE/S@10, Target Only remains strongest for Place FDE/S@10, and No Prompt retains the largest Pick--Place gap. The agreement with the all-episode analysis indicates that the role-specific result is not driven by low-confidence contact boundaries.

\section{Statistical Details for Prompting Baselines}
Table~\ref{tab:baseline-confidence-intervals} reports the paired-bootstrap confidence intervals corresponding to the three-seed results in Table~\ref{tab:baseline-results}.
\begin{table*}[t]
  \centering
  \caption{Hierarchical paired-bootstrap statistics for the prompting baselines, using 10,000 resamples over seeds and validation episodes. Each row reports BBox+Visual minus the named baseline with a 95\% confidence interval. Success-rate differences are in percentage points (pp); bold intervals exclude zero.}
  \label{tab:baseline-confidence-intervals}
  \small
  \setlength{\tabcolsep}{5pt}
  \resizebox{0.72\textwidth}{!}{%
    \begin{tabular}{lcc}
      \toprule
      Comparison & $\Delta$Pos. L2 (cm) [95\% CI] & $\Delta$FDE (cm) [95\% CI] \\
      \midrule
      vs. No prompt   & \textbf{$-2.4905$ [$-2.8160,-2.1132$]} & \textbf{$-6.6132$ [$-7.3008,-5.8663$]} \\
      vs. Target only & \textbf{$-0.1683$ [$-0.3084,-0.0150$]} & $-0.4327$ [$-0.8186,0.0368$] \\
      vs. Visual      & $+0.0774$ [$-0.0699,0.2009$] & $-0.1097$ [$-0.4149,0.1794$] \\
      vs. BBox        & \textbf{$-0.2466$ [$-0.4217,-0.0560$]} & \textbf{$-0.7069$ [$-1.0822,-0.2920$]} \\
      \bottomrule
    \end{tabular}%
  }
  \vspace{3pt}

  \resizebox{0.92\textwidth}{!}{%
    \begin{tabular}{lccc}
      \toprule
      Comparison & $\Delta$S@2 (pp) [95\% CI] & $\Delta$S@5 (pp) [95\% CI] & $\Delta$S@10 (pp) [95\% CI] \\
      \midrule
      vs. No prompt   & $-0.51$ [$-1.61,0.43$] & \textbf{$+1.84$ [$0.14,3.34$]} & \textbf{$+13.95$ [$12.21,15.65$]} \\
      vs. Target only & $+0.46$ [$-0.05,0.95$] & \textbf{$+1.14$ [$0.22,2.07$]} & \textbf{$+3.00$ [$1.43,4.53$]} \\
      vs. Visual      & $-0.19$ [$-0.70,0.39$] & \textbf{$-1.35$ [$-2.25,-0.46$]} & $+0.20$ [$-2.33,3.22$] \\
      vs. BBox        & \textbf{$+0.81$ [$0.16,1.43$]} & \textbf{$+2.29$ [$1.16,3.46$]} & \textbf{$+4.12$ [$2.56,5.74$]} \\
      \bottomrule
    \end{tabular}%
  }
\end{table*}

\section{Additional Prompt Analyses}
Table~\ref{tab:prompt-noise} gives the complete prompt-noise results summarized in the main paper. Center perturbations cause monotonic degradation, whereas moderate box scaling has little effect. Missing-target performance is substantially worse than missing-object performance.

\begin{table}[t]
  \centering
  \caption{Prompt-noise robustness on all 10,024 validation samples. Combined values are offline endpoint criteria at 5/10 cm.}
  \label{tab:prompt-noise}
  \scriptsize
  \setlength{\tabcolsep}{3pt}
  \resizebox{\columnwidth}{!}{%
    \begin{tabular}{lcccc}
      \toprule
      Condition & FDE (cm) & $\Delta$FDE (\%) & Rot. geo. ($^\circ$) & Combined@5/10 (\%) \\
      \midrule
      Correct prompt & 11.4655 & -- & 9.95 & 15.81 / 38.67 \\
      Jitter 2\%      & 11.6984 & +2.03 & 10.20 & 15.65 / 38.06 \\
      Jitter 5\%      & 12.3716 & +7.90 & 10.27 & 14.66 / 35.01 \\
      Jitter 10\%     & 14.2021 & +23.87 & 10.89 & 11.40 / 27.86 \\
      Jitter 20\%     & 17.5195 & +52.80 & 11.28 & 9.13 / 24.95 \\
      Scale 0.8       & \textbf{11.3393} & -1.10 & \textbf{9.77} & \textbf{16.31 / 38.93} \\
      Scale 0.9       & 11.4147 & -0.44 & 9.84 & 15.89 / 38.84 \\
      Scale 1.1       & 11.5588 & +0.81 & 10.02 & 15.73 / 37.95 \\
      Scale 1.2       & 11.6245 & +1.39 & 10.20 & 15.17 / 36.88 \\
      Missing object  & 13.2412 & +15.49 & 10.19 & 12.89 / 31.74 \\
      Missing target  & 18.2588 & +59.25 & 10.78 & 12.51 / 29.14 \\
      \bottomrule
    \end{tabular}%
  }
\end{table}

Table~\ref{tab:prompt-noise} shows a clear dose--response relationship for prompt-center error. Relative to the correct-prompt FDE of 11.47 cm, 2\% jitter changes FDE by only +2.03\%, but 5\%, 10\%, and 20\% jitter increase it by 7.90\%, 23.87\%, and 52.80\%, respectively. Combined@10 correspondingly falls from 38.67\% to 38.06\%, 35.01\%, 27.86\%, and 24.95\%. SPOT is therefore tolerant to small annotation displacement but degrades predictably once the prompt center moves substantially.

Changing box scale from $0.8\times$ to $1.2\times$ has a much smaller effect: FDE remains within $-1.10\%$ to $+1.39\%$ of the correct-prompt value, and Combined@10 remains between 36.88\% and 38.93\%. The small improvement at $0.8\times$ is a single-seed fluctuation and is not evidence that shrinking the box is beneficial. Together with the jitter trend, this indicates that prompt location is more important than precise box extent. Finally, removing the target increases FDE by 59.25\%, compared with 15.49\% when removing the object, and reduces Combined@10 to 29.14\% rather than 31.74\%. Rotation changes are comparatively modest (mean Rot.~geo. stays between $9.77^\circ$ and $11.28^\circ$ across all perturbations), so prompt corruption primarily redirects translation instead of uniformly degrading pose prediction.

\begin{table*}[t]
  \centering
  \caption{Additional seed-42 task-oriented metrics on all 10,024 validation samples. Rot.~geo. reports mean/endpoint geodesic error in degrees; R@10/20 is endpoint rotation success; G-end is endpoint gripper-state accuracy; and C@5/10 is combined offline endpoint success. Grip.~L1 is in millimeters. Correct prompt variants use their corresponding trained checkpoints, whereas the shuffled and role-swap rows are inference-time interventions on the BBox+Visual checkpoint.}
  \label{tab:additional-prompt-metrics}
  \scriptsize
  \setlength{\tabcolsep}{3.2pt}
  \resizebox{0.98\textwidth}{!}{%
    \begin{tabular}{lcccccc}
      \toprule
      Condition & FDE (cm) & Rot.~geo. mean/end ($^\circ$) & Grip.~L1 (mm) & R@10/20 (\%) & G-end (\%) & C@5/10 (\%) \\
      \midrule
      No prompt       & 17.39 & 10.10/16.30 & 10.98 & 36.69/71.29 & 80.34 & 16.02/29.94 \\
      Object only     & 17.28 & 10.26/16.18 & 10.96 & 36.95/72.15 & 81.09 & 15.60/31.57 \\
      Target only     & 11.44 & 10.47/16.79 & 12.05 & 31.70/72.40 & 79.77 & 15.01/34.88 \\
      Point prompt    & 11.38 & 9.94/16.21  & \textbf{10.61} & 34.61/73.57 & \textbf{82.77} & 16.05/37.01 \\
      BBox prompt     & 11.76 & 10.25/16.23 & 11.73 & 34.61/72.96 & 80.79 & 14.43/34.47 \\
      Visual prompt   & 11.75 & \textbf{9.42/14.80} & 10.95 & \textbf{39.04/77.89} & 80.87 & \textbf{16.97}/37.29 \\
      BBox + visual   & \textbf{11.47} & 9.95/15.53 & 11.33 & 35.48/77.26 & 81.04 & 15.81/\textbf{38.67} \\
      Shuffled object & 13.84 & 10.90/17.58 & 11.48 & 30.18/71.43 & 80.56 & 12.68/29.93 \\
      Shuffled target & 18.47 & 11.09/17.93 & 11.83 & 33.10/69.28 & 79.34 & 11.76/26.68 \\
      Role swap       & 22.08 & 11.23/17.85 & 11.59 & 30.94/68.86 & 79.62 & 9.86/24.77 \\
      \bottomrule
    \end{tabular}%
  }
\end{table*}

Table~\ref{tab:additional-prompt-metrics} extends the prompt comparison beyond translation error. Object Only remains close to No Prompt, whereas Target Only explains most of the endpoint gain. Visual Prompt gives the best rotation and Combined@5 results, while BBox+Visual gives the best Combined@10 result. Prompt corruption affects FDE and combined success much more strongly than endpoint gripper state, supporting a spatial redirection effect rather than a uniform degradation of every control variable.

\section{Additional Architecture Ablations}
\par\medskip
\noindent\begin{minipage}{\columnwidth}
  \centering
  \captionof{table}{Ablation on visual encoder type. We compare frozen base-size visual encoders under the same SPOT configuration. Translation errors are in centimeters and Grip.~L1 is in millimeters; lower is better.}
  \label{tab:ablation-vision-encoder-type}
  \scriptsize
  \setlength{\tabcolsep}{2pt}
  \resizebox{\columnwidth}{!}{%
    \begin{tabular}{clcccc}
      \toprule
      Split & Encoder & FDE (cm) & Pos. L2 (cm) & Rot. geo. ($^\circ$) & Grip. L1 (mm) \\
      \midrule
      \multirow{5}{*}{All} & PE-Base & 15.44 & 8.91 & 10.91 & 10.85 \\
      & SigLIP-Base & 12.04 & 7.24 & \textbf{9.58} & \textbf{9.21} \\
      & SAM-Base & 17.59 & 9.29 & 10.93 & 9.28 \\
      & EVA2-Base & 14.22 & 8.26 & 10.51 & 10.50 \\
      & DINOv2-Base & \textbf{11.47} & \textbf{6.99} & 9.95 & 11.33 \\
      \midrule
      \multirow{5}{*}{Scene 1} & PE-Base & 11.37 & 6.87 & 7.70 & 7.49 \\
      & SigLIP-Base & \textbf{10.83} & \textbf{6.64} & \textbf{7.38} & \textbf{6.74} \\
      & SAM-Base & 14.18 & 7.52 & 7.81 & 6.76 \\
      & EVA2-Base & 10.96 & 6.65 & 7.64 & 7.28 \\
      & DINOv2-Base & 11.23 & 7.00 & 7.67 & 6.93 \\
      \midrule
      \multirow{5}{*}{Scene 2} & PE-Base & 18.88 & 10.19 & 12.42 & 9.67 \\
      & SigLIP-Base & 12.48 & 7.02 & \textbf{10.43} & \textbf{8.58} \\
      & SAM-Base & 19.98 & 10.01 & 12.23 & 10.17 \\
      & EVA2-Base & 15.64 & 8.61 & 11.40 & 9.52 \\
      & DINOv2-Base & \textbf{11.16} & \textbf{6.32} & 10.93 & 9.75 \\
      \midrule
      \multirow{5}{*}{Scene 3} & PE-Base & 18.22 & 10.83 & 14.46 & 17.88 \\
      & SigLIP-Base & 13.61 & 8.53 & \textbf{12.25} & 13.99 \\
      & SAM-Base & 20.90 & 11.54 & 14.56 & \textbf{12.54} \\
      & EVA2-Base & 17.96 & 10.55 & 14.21 & 17.00 \\
      & DINOv2-Base & \textbf{12.24} & \textbf{7.79} & 12.58 & 20.48 \\
      \bottomrule
    \end{tabular}%
  }
\end{minipage}
\par\medskip

Table~\ref{tab:ablation-vision-encoder-type} exposes a metric-dependent tradeoff rather than a universal winner. DINOv2 has the best overall FDE/Pos.~L2 (11.47/6.99 cm), whereas SigLIP has the best Rot.~geo./Grip.~L1 ($9.58^\circ$/9.21 mm). SAM reinforces this decoupling: it has the worst FDE but nearly matches SigLIP on gripper error. Encoder differences also grow with scene complexity: the FDE spread increases from 3.35 cm in Scene 1 to 8.66 cm in Scene 3. DINOv2 leads translation in Scenes 2/3, while SigLIP leads their rotation errors; no encoder dominates every control variable. We therefore choose DINOv2 for its spatial-trajectory accuracy, not because it is uniformly best, and view vision pretraining objective as a source of complementary localization and pose-control biases.

\section{Additional Ablation: Vision Encoder Size}
We further compare DINOv2 backbones of different sizes under the same SPOT setting. All variants use the combined visual and coordinate bounding-box prompt, frozen visual features, flow matching, and the same scene-aware evaluation protocol. Table~\ref{tab:ablation-vision-encoder-size} reports the results.

\begin{table}[t]
  \centering
  \caption{Ablation on DINOv2 encoder size. Translation errors are in centimeters and Grip.~L1 is in millimeters; lower is better.}
  \label{tab:ablation-vision-encoder-size}
  \small
  \setlength{\tabcolsep}{2pt}
  \resizebox{\columnwidth}{!}{%
  \begin{tabular}{clcccc}
    \toprule
    Split & Encoder & FDE (cm) & Pos. L2 (cm) & Rot. geo. ($^\circ$) & Grip. L1 (mm) \\
    \midrule
    \multirow{3}{*}{All}
      & DINOv2-Small & \textbf{11.29} & 7.01 & 10.70 & 13.86 \\
      & DINOv2-Base & 11.47 & 6.99 & 9.95 & 11.33 \\
      & DINOv2-Large & 11.35 & \textbf{6.79} & \textbf{9.39} & \textbf{10.81} \\
    \midrule
    \multirow{3}{*}{Scene 1}
      & DINOv2-Small & 10.85 & 6.85 & 8.15 & 7.19 \\
      & DINOv2-Base & 11.23 & 7.00 & 7.67 & 6.93 \\
      & DINOv2-Large & \textbf{9.88} & \textbf{6.09} & \textbf{7.18} & \textbf{6.76} \\
    \midrule
    \multirow{3}{*}{Scene 2}
      & DINOv2-Small & \textbf{10.57} & \textbf{6.17} & 11.48 & \textbf{8.85} \\
      & DINOv2-Base & 11.16 & 6.32 & 10.93 & 9.75 \\
      & DINOv2-Large & 12.44 & 6.79 & \textbf{10.45} & 9.78 \\
    \midrule
    \multirow{3}{*}{Scene 3}
      & DINOv2-Small & 12.82 & 8.29 & 14.04 & 30.91 \\
      & DINOv2-Base & \textbf{12.24} & \textbf{7.79} & 12.58 & 20.48 \\
      & DINOv2-Large & 12.51 & 7.96 & \textbf{11.81} & \textbf{18.71} \\
    \bottomrule
  \end{tabular}%
  }
\end{table}

Increasing DINOv2 size mainly improves orientation and gripper prediction, while endpoint accuracy saturates: the overall FDE spread is only 0.18 cm and Small is marginally best. Large improves Rot.~geo. and Grip.~L1 over Base by $0.56^\circ$ and 0.52 mm, but its Scene-2 FDE is 1.87 cm worse than Small. Capacity is more useful for non-translation variables in the hardest scene, where Large reduces gripper error by 12.20 mm relative to Small, yet Base still has the best Scene-3 FDE/Pos.~L2. These reversals rule out a simple scaling trend and suggest that backbone capacity and visual domain interact. We retain Base because it balances translation, pose/gripper accuracy, and compute, rather than selecting a different size for each scene.

\section{More Visualization Results}
\begin{figure}[t]
    \centering
    \includegraphics[width=0.98\linewidth]{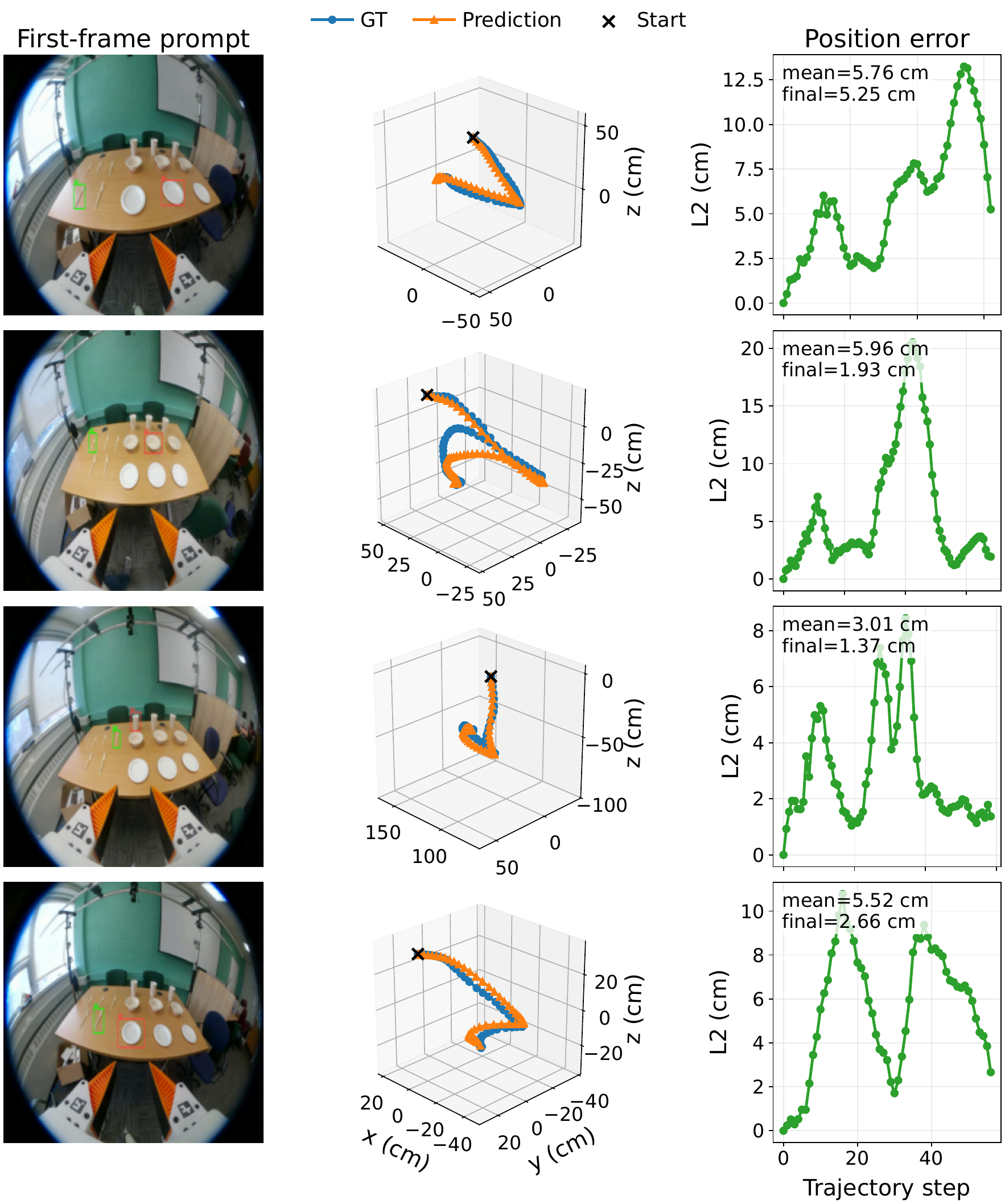}
    \caption{\textbf{Additional full-trajectory visualizations for Scene 1.} Stitched trajectories and position errors are reported in centimeters.}
    \label{fig:scene1-vis}
\end{figure}

\begin{figure}[t]
    \centering
    \includegraphics[width=0.98\linewidth]{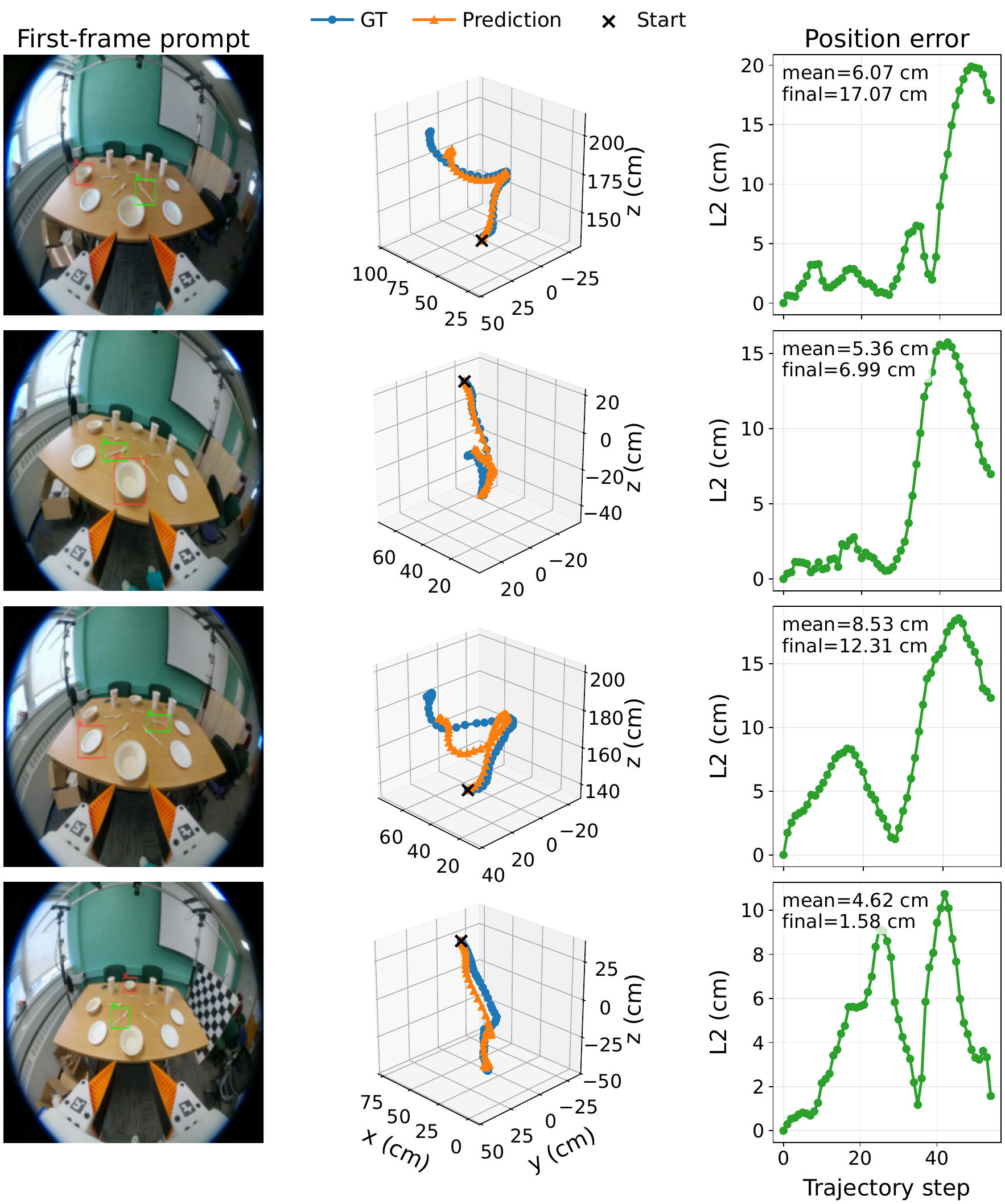}
    \caption{\textbf{Additional full-trajectory visualizations for Scene 2.} Stitched trajectories and position errors are reported in centimeters.}
    \label{fig:scene2-vis}
\end{figure}

\begin{figure}[t]
    \centering
    \includegraphics[width=0.98\linewidth]{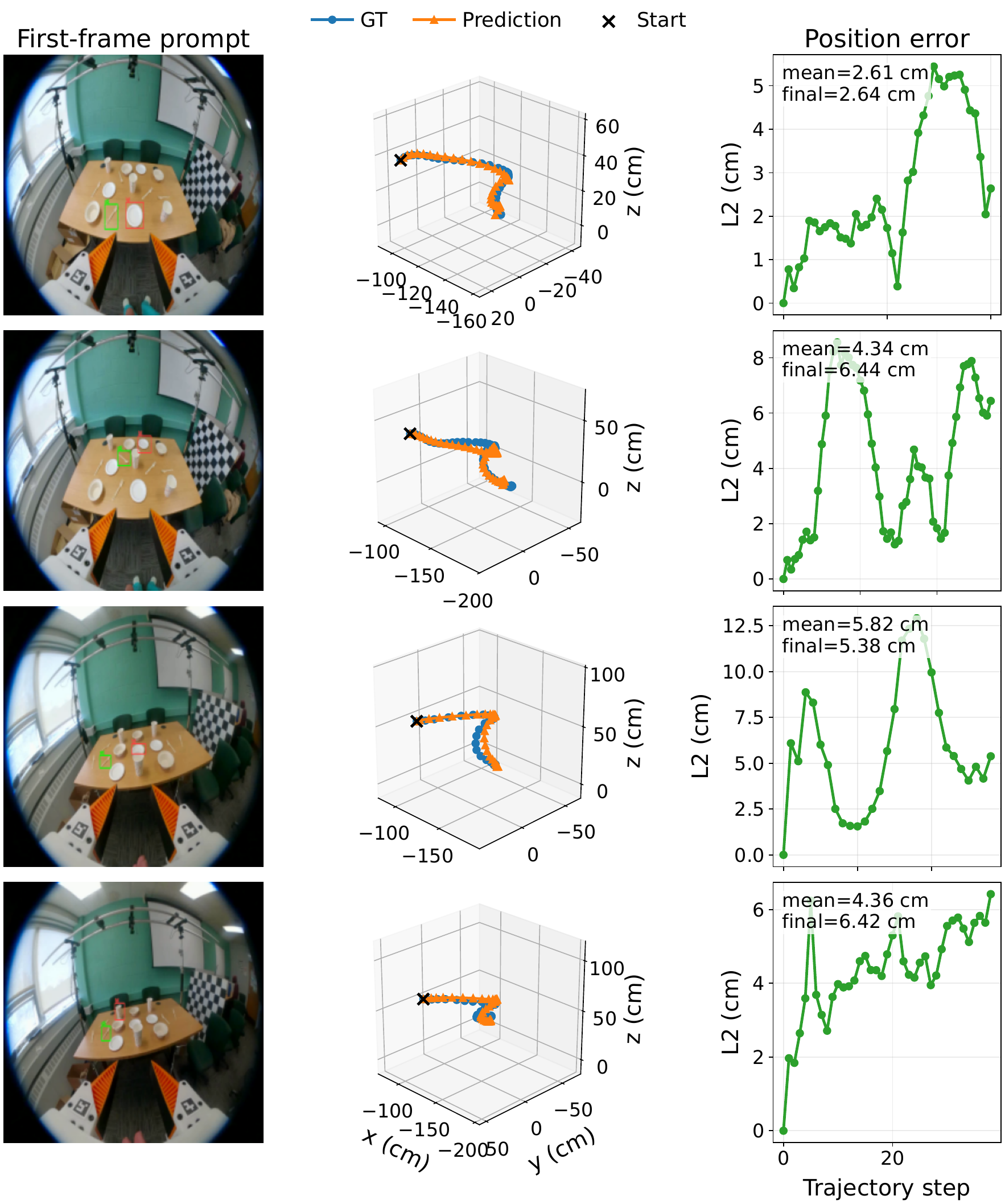}
    \caption{\textbf{Additional full-trajectory visualizations for Scene 3.} Stitched trajectories and position errors are reported in centimeters.}
    \label{fig:scene3-vis}
\end{figure}

\begin{figure}
    \centering
    \includegraphics[width=0.85\linewidth]{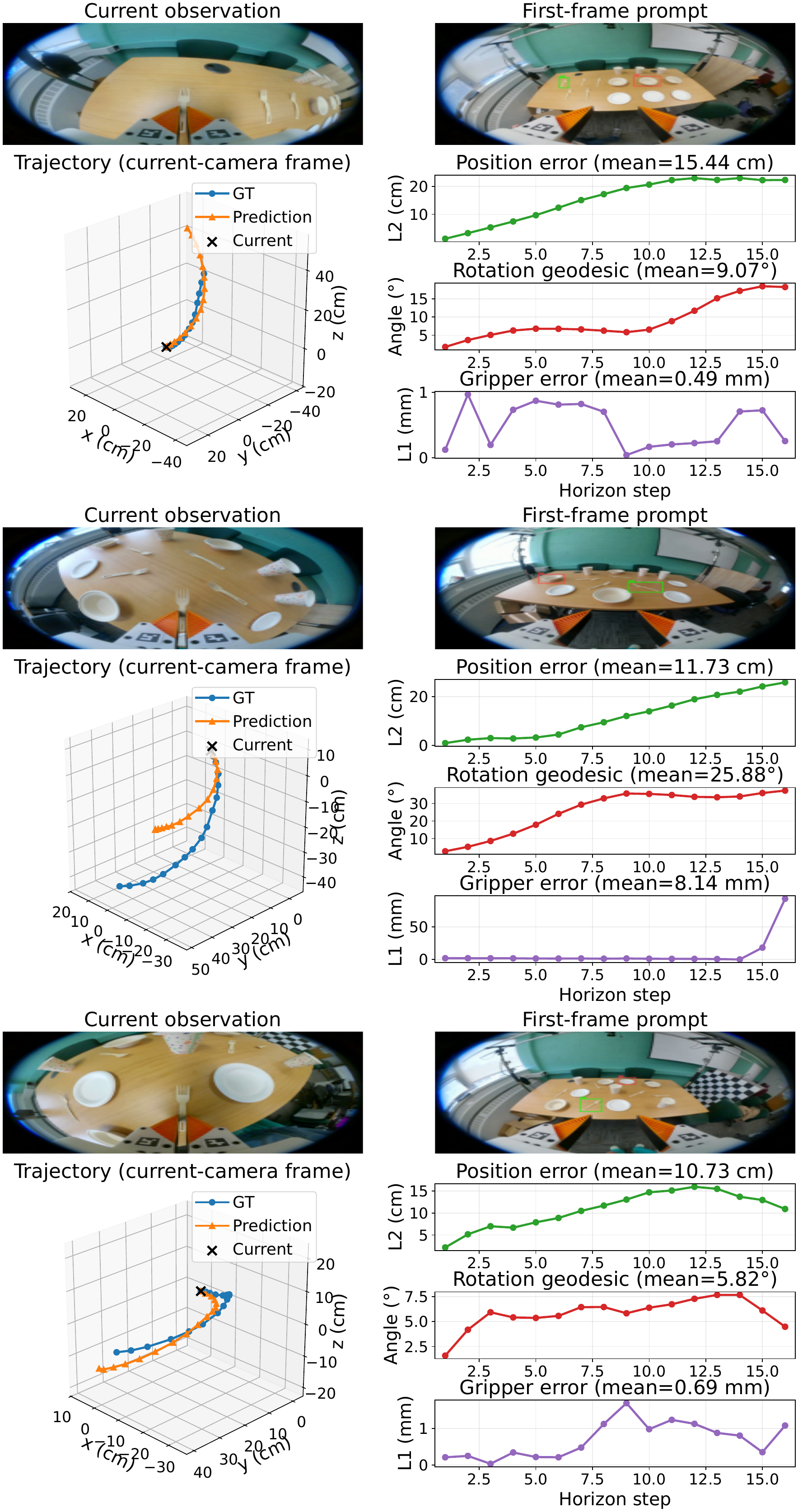}
    \caption{\textbf{Additional first-chunk visualizations for three scenes.} Scenes 1--3 are shown top to bottom; units are cm (trajectory/position), degrees (rotation geodesic), and mm (gripper width).}
    \label{fig:first-chunk-vis}
\end{figure}

We provide additional full-episode trajectories from the three scenes in Figs.~\ref{fig:scene1-vis}, \ref{fig:scene2-vis}, and \ref{fig:scene3-vis}, as well as further first-chunk examples in Fig.~\ref{fig:first-chunk-vis}.

\section{Limitations}
SP-VTP is evaluated as open-loop trajectory prediction rather than closed-loop robot control. Although stitched trajectories indicate that SPOT preserves the coarse episode structure, local chunk errors can accumulate over long horizons, especially under large camera and EE motion. The current dataset focuses on fork pick-and-place tasks with table-top receptacles, so the scope of the claims is limited to spatially prompted egocentric manipulation in this task family. Broader objects, deformable items, multi-step tasks, and real-time closed-loop execution remain important directions for future work.

The method also assumes that the first-frame object and target prompts are available and reasonably accurate. Strong prompt noise, severe occlusion, or target regions that leave the camera view may reduce performance. Finally, while frozen foundation visual encoders perform better in our experiments, their behavior may depend on the visual domain, camera viewpoint, and object categories.

\section{Ethics and Broader Impact}
EgoSPT is collected and annotated by trained experts rather than crowdsourced workers. The videos are captured from an egocentric device during table-top manipulation and are intended to avoid collecting identifiable personal information. The dataset is designed for research on visually grounded manipulation and does not involve high-risk generated media, scraped web data, or personal decision making.

Spatial prompting can make robot task specification more direct and accessible, especially in cluttered scenes where language instructions are ambiguous. At the same time, improved manipulation policies may be unsafe if deployed without closed-loop monitoring, collision checking, or task-level constraints. Any deployment should therefore include safety checks, workspace limits, and human supervision appropriate to the robot platform and environment.

\end{document}